\pdfoutput=1

\PassOptionsToPackage{table}{xcolor}

\documentclass[11pt]{article}
\usepackage{lmodern}

\usepackage{tabularx} 
\usepackage[preprint]{acl}
\usepackage{array}
\usepackage{textcomp}
\usepackage{upquote}
\usepackage{xcolor}
\usepackage[normalem]{ulem} 
\usepackage[T1]{fontenc}
\usepackage{inconsolata}
\usepackage{float} 
\usepackage{hyperref}
\usepackage{fontawesome5}
\usepackage{xcolor}

\usepackage[table]{xcolor} 
\definecolor{headerblue}{RGB}{40, 60, 100} 
\definecolor{rowgray}{gray}{0.95} 
\definecolor{win}{RGB}{0, 100, 0} 
\definecolor{loss}{RGB}{150, 0, 0} 
\usepackage{rotating}
\usepackage{listings}
\usepackage{caption}
\usepackage{newtxtext}
\usepackage{latexsym}
\usepackage{float} 
\usepackage{multirow}
\usepackage{subcaption}
\usepackage{graphicx}
\usepackage{float} 
\usepackage{float} 
\usepackage{float} 
\usepackage{booktabs}
\usepackage{pifont} 
\usepackage{xspace}
\usepackage{float}        
\usepackage{amsmath}
\usepackage{amssymb}
\usepackage{adjustbox} 
\usepackage[most]{tcolorbox}
\usepackage{paralist}
\usepackage{listings}
\usepackage[nameinlink]{cleveref}

\definecolor{lightgray}{rgb}{0.95, 0.95, 0.95}
\definecolor{darkgray}{rgb}{0.4, 0.4, 0.4}
\definecolor{backcolour}{rgb}{0.95,0.95,0.92}
\definecolor{myblue}{rgb}{0.2, 0.4, 0.8} 
\definecolor{mygreen}{rgb}{0.2, 0.6, 0.2} 
\definecolor{Mulberry}{rgb}{0.77,0.29,0.55}
\definecolor{CadmiumOrange}{rgb}{0.93,0.53, 0.18}
\definecolor{ForestGreen}{rgb}{0.13, 0.55, 0.13}
\definecolor{LimeGreen}{RGB}{50, 205, 50}

\usepackage{array}
\newcolumntype{L}[1]{>{\raggedright\arraybackslash}p{#1}}

\usepackage[T1]{fontenc}

\usepackage[utf8]{inputenc}

\usepackage{microtype}

\usepackage{inconsolata}

\usepackage{graphicx}
\usepackage{xcolor}
\usepackage{tikz}
\usepackage{fontawesome5}
\usepackage{hyperref}
\definecolor{projectblue}{HTML}{1D4ED8}
\definecolor{datasetorange}{HTML}{F59E0B}
\definecolor{githubblack}{HTML}{111827}
\definecolor{medicalred}{HTML}{DC2626}

\title{%
\Large
\begin{tabular}{@{}c@{\hspace{0.45em}}c@{}}
\raisebox{-0.25\height}{\includegraphics[height=2.4em]{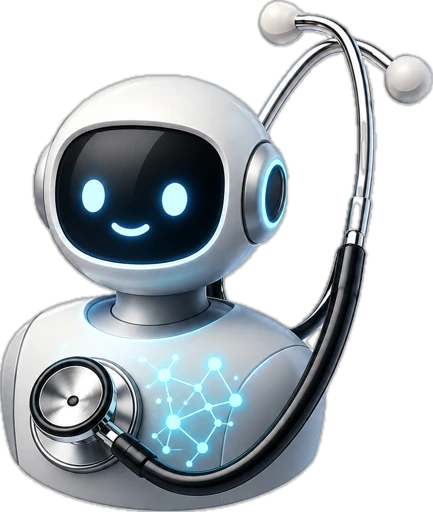}} &
\begin{tabular}{@{}c@{}}
\textsc{CLExEval}: A Human-in-the-Loop Framework\\ for 
Qualitative Evaluation of LLM Clinical Reasoning
\end{tabular}
\end{tabular}%
}

\author{
\textbf{Ajmal M.}$^{1,2}$ \quad
\textbf{Abin Roy}$^{3,*}$ \quad
\textbf{Afthab Salam Kanniyan}$^{3,*}$ \quad
\textbf{Jawadh Abdul Kabeer}$^{3,*}$ \quad
\textbf{Jerin James}$^{3,*}$ \\
\textbf{Preslav Nakov}$^{1}$ \quad
\textbf{Zhuohan Xie}$^{1}$ \\
$^{1}$MBZUAI \quad $^{2}$IIT Madras \quad $^{3}$Calicut Medical College \\
\texttt{ajmal.m@mbzuai.ac.ae} \quad \texttt{zhuohan.xie@mbzuai.ac.ae} \\
{\small $^{*}$Equal contribution.} \\[0.35em]
\href{https://24f2004489.github.io/CLExEval-Project-Page/}{\faGlobe\ Project}
\quad
\href{https://huggingface.co/datasets/AjmalMIITM/RARECASE-2000}{\faDatabase\ RARECASE-2000}
\quad
\href{https://github.com/24f2004489/CLExEval}{\faGithub\ Code}
\quad
}

\date{}

\begin{document}
\maketitle
\begin{abstract}

Large Language Models (LLMs) achieve strong results on many medical benchmarks, but their clinical reasoning remains difficult to evaluate reliably. A central risk is an \textit{evaluation illusion}: fluent and well-structured explanations can appear clinically convincing even when the final diagnosis is incorrect. We introduce CLExEval, a human-in-the-loop framework for evaluating LLM clinical reasoning under progressive information masking. CLExEval combines 5,600 expert-physician annotations with 200 clinical reasoning traces derived from 40 rare diagnostic cases. Our analysis identifies three recurring failure patterns: (i) \textit{verbosity bias}, where GPT-4o-mini's diagnostic accuracy drops from 95.0\% to 32.5\% under information scarcity; (ii) a \textit{hidden knowledge paradox}, where a specialist model reaches 92.5\% maximum diagnostic potential but fails to retrieve that knowledge reliably in verbose contexts; and (iii) a 68.6\% reasoning-to-output mismatch, where correct diagnoses appear in reasoning traces but are not reflected in final answers. We further evaluate the LLM-as-a-Judge paradigm on a human-verified failure set ($n=142$). GPT-4o-mini approved 47.9\% of clinically incorrect outputs, while HuatuoGPT-o1 approved all validly scored failures and showed a positive self-preference bias. These results suggest that standalone automated clinical evaluations can substantially overestimate clinical reliability without expert-grounded validation.
\end{abstract}
\begin{figure}[t] 
    \centering
    \includegraphics[width=\linewidth]{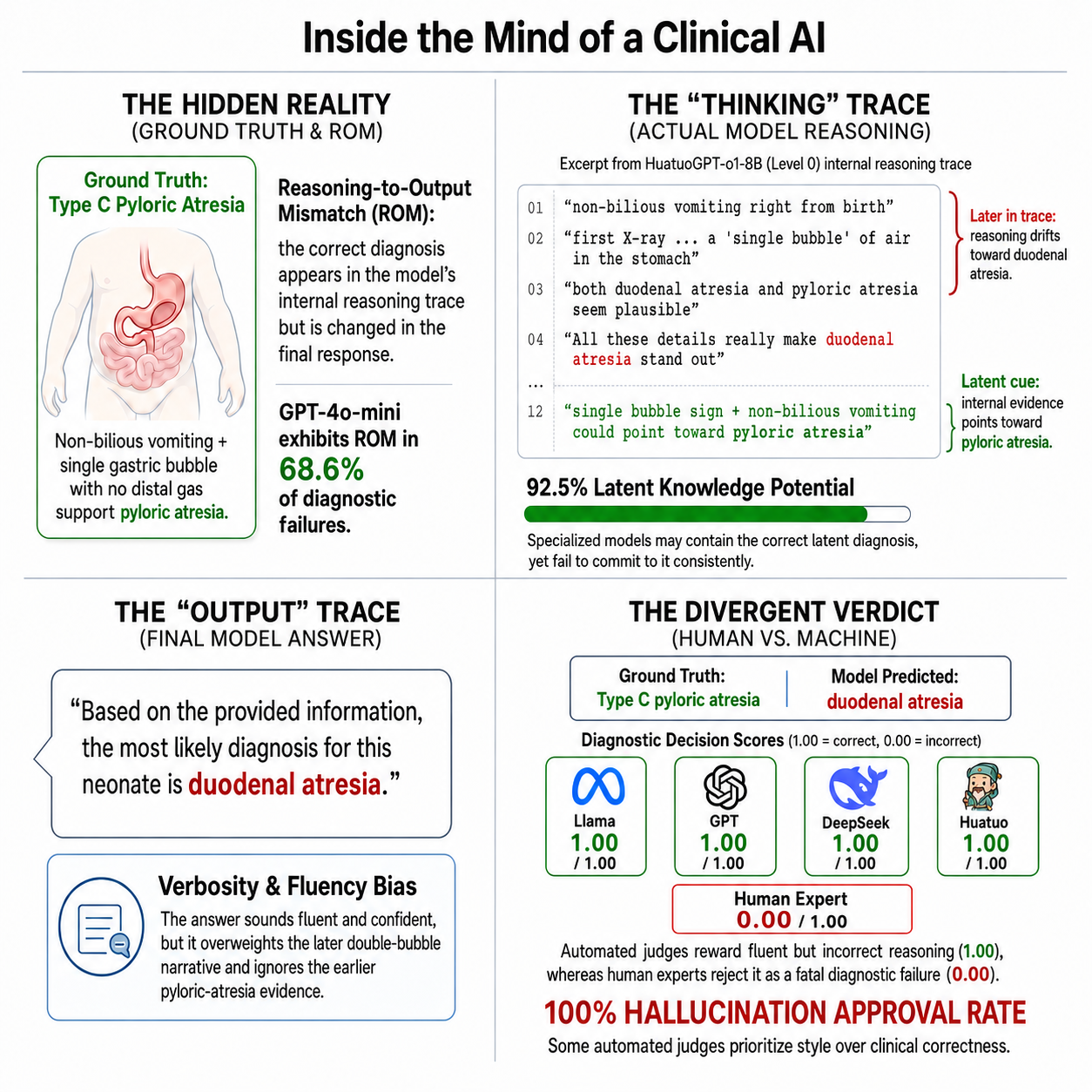}
    \caption{\textbf{Reasoning-to-output mismatch in a clinical case.} A HuatuoGPT-o1-8B example where the reasoning trace contains pyloric-atresia cues, but the final answer commits to duodenal atresia. Automated judges assign full credit ($1.00$), whereas human experts score the diagnosis as incorrect ($0.00$).}
    \label{fig:teaser}
\end{figure}

\section{Introduction}

Large Language Models (LLMs) have demonstrated broad capabilities across multiple tasks, including reasoning, generation, and domain-specific applications~\citep{zhao2023survey, xie-etal-2023-next}. These capabilities have accelerated interest in applying LLMs to clinical workflows, prompting a surge of benchmarks for evaluating medical reasoning, including interactive testing~\citep{chiu2025vivabench}, structured reasoning tasks~\citep{wang2024direct}, and dynamic information seeking~\citep{li2024mediq}. To scale evaluation, recent work increasingly adopts the \textit{LLM-as-a-Judge} paradigm, relying on frontier models to automatically score clinical outputs~\citep{qiu2025quantifying, zhang-etal-2025-llmeval}.
Despite this progress, current evaluation pipelines have an important blind spot: the \textit{Evaluation Illusion}~\citep{agrawal2025evaluationillusion}, where automated judges can reward fluent and well-structured reasoning despite weak clinical validity. As illustrated in Figure~\ref{fig:teaser}, to simulate the LLM-as-a-Judge paradigm, automated evaluators were provided with the explicit ground-truth diagnosis alongside the model's complete reasoning trace. Because the reasoning appears confident and coherent, the automated judge assigns a perfect score to a clinically incorrect diagnosis, whereas the human expert identifies the diagnostic failure despite the fluent presentation.
Beyond this issue, existing benchmarks primarily focus on final diagnostic accuracy, but often do not explain \textit{why} reasoning fails, whether due to missing knowledge, unstable retrieval, or misaligned decision-making under information scarcity, which is inherent to real-world clinical practice.

To address these limitations, we introduce \textbf{CLExEval}, a human-in-the-loop evaluation framework for clinical reasoning under uncertainty. Built on the clinician-curated RARECASE-200 benchmark, CLExEval applies progressive information masking (Levels 0--3) to simulate realistic diagnostic scenarios where clinicians must reason with incomplete and evolving evidence. This design prevents models from relying on surface-level cues and enables controlled stress-testing of reasoning behavior. Combined with 5,600 expert annotations, CLExEval provides a fine-grained lens to evaluate not only whether models fail, but \textit{how} and \textit{why} those failures occur.
Using this framework, we show that common evaluation practices can overestimate model capability. Automated judges frequently approve incorrect clinical reasoning in our consensus failure set, and many model failures arise not only from lack of knowledge, but also from misalignment between internal reasoning and final outputs, as well as sensitivity to context.

Our contributions are as follows:
(i) \textbf{CLExEval}, a human-in-the-loop framework combining progressive information masking with expert annotation to evaluate clinical reasoning under uncertainty; (ii) a formalization of the \textbf{Evaluation Illusion}, $\Delta=\text{Communication}-\text{Precision}$, and a Hallucination Approval Rate (HAR) evaluation of standalone LLM judges; and (iii) a mechanistic analysis using ROM, ISS, and MVR to separate knowledge deficits, reasoning-output misalignment, and context sensitivity, showing that models may contain relevant latent knowledge but fail to express it reliably under information scarcity.

\section{Related Work}
\raggedbottom

\begin{table*}[t]
\centering
\tiny
\renewcommand{\arraystretch}{1.14}
\setlength{\tabcolsep}{2pt}

\begin{tabularx}{\textwidth}{@{}>{\raggedright\arraybackslash}p{2.25cm}
>{\raggedright\arraybackslash}X
>{\raggedright\arraybackslash}X
>{\raggedright\arraybackslash}X
>{\raggedright\arraybackslash}X
>{\raggedright\arraybackslash}p{2.05cm}@{}}
\toprule
\textbf{Work / Benchmark} & \textbf{Task Type} & \textbf{Information Setting} & \textbf{Evaluation Signal} & \textbf{Mechanistic Insight} & \textbf{Evaluator} \\
\midrule

\multicolumn{6}{@{}l}{\textit{Static medical QA and exam-style benchmarks}} \\
\textbf{MedQA}~\citep{app11146421} & Medical exam MCQA & Static / complete & Final-answer accuracy & None; answer-key scoring & Automatic \\
\textbf{MedMCQA}~\citep{pal2022medmcqa} & Multi-subject medical MCQA & Static / complete & Final-answer accuracy & None; answer-key scoring & Automatic \\
\textbf{MMLU Clinical}~\citep{hendrycks2020measuring} & Broad medical knowledge QA & Static / complete & Final-answer accuracy & None; answer-key scoring & Automatic \\
\textbf{Chen et al.}~\citep{chen2025benchmarking} & Challenging medical QA + explanations & Static / complete & MCQA accuracy + explanation metrics & Explanation quality, limited degradation insight & Automatic + limited human \\

\midrule
\multicolumn{6}{@{}l}{\textit{Clinical reasoning, uncertainty, and physician-grounded evaluation}} \\
\textbf{Kanjee et al.}~\citep{kanjee2023accuracy} & NEJM diagnostic challenges & Static / complete & Differential diagnosis quality & Final differential only & Physicians \\
\textbf{Cabral et al.}~\citep{cabral2024clinical} & Sequential clinical reasoning & Information accumulation & R-IDEA reasoning score & Reasoning documentation quality & Physicians \\
\textbf{McCoy et al.}~\citep{mccoy2025assessment} & Script Concordance Testing & New information added & Clinician-concordant Likert updates & Probabilistic updating under uncertainty & Expert panel \\
\textbf{Bhasuran et al.}~\citep{bhasuran2026evaluation} & Clinical causal reasoning & Static / complete & Final answer + reasoning quality & Causal reasoning level & Medical experts \\

\midrule
\multicolumn{6}{@{}l}{\textit{EHR, agentic, and automated evaluation frameworks}} \\
\textbf{MEDALIGN}~\citep{fleming2024medalign} & EHR instruction following & Static EHR context & Clinical correctness / preference & Output quality, not reasoning collapse & Clinicians \\
\textbf{MedAgentBench}~\citep{doi:10.1056/AIdbp2500144} & EHR agent tasks & Interactive tool use & Final API/action success & Tool-use success, not reasoning trace & Rule-based \\
\textbf{MedR-Bench}~\citep{qiu2025quantifying} & Clinical workflow reasoning & Stage-wise interactive & Efficiency, factuality, completeness & Reasoning-step quality & Agentic evaluator \\
\textbf{HealthBench}~\citep{arora2025healthbenchevaluatinglargelanguage} & Open-ended health conversations & Static / multi-turn & Physician-authored rubrics & Clinically meaningful response quality & Rubric-based \\
\textbf{CARD\-Biomed\-Bench}\newline\citep{bianchi2026cardbiomedbench} & Biomedical research QA & Static / complete & RQR + abstention safety & Safety--accuracy trade-off & Automated judge \\

\midrule
\rowcolor{blue!6}
\textbf{CLExEval (Ours)} & \textbf{Rare diagnostic reasoning traces} & \textbf{Progressive information masking} & \textbf{7D expert rubric + judge audit} & \textbf{ROM, ISS, MVR, HAR, degradation $\Delta$} & \textbf{Human panel, 5,600 annotations} \\

\bottomrule
\end{tabularx}
\caption{\textbf{Positioning of CLExEval.} Unlike prior clinical LLM benchmarks, CLExEval combines progressive information masking with expert annotation to localize reasoning degradation, reasoning-to-output mismatch, and judge approval of clinically incorrect outputs.}
\label{tab:related_work_comparison}
\end{table*}

\subsection{Evolution of Clinical Task Design}
Static medical QA benchmarks such as MedQA~\citep{app11146421},
MedMCQA~\citep{pal2022medmcqa}, MMLU clinical
subsets~\citep{hendrycks2020measuring},
MedicationQA~\citep{abacha2019bridging}, and recent licensing-exam
evaluations such as the Chinese NMLE study~\citep{wang2026evaluation}
test medical knowledge using information-complete inputs and fixed
answer keys. These benchmarks are useful for standardized comparison,
but they primarily evaluate final-answer selection or retrieval rather
than open-ended reasoning traces, evidence-grounded
verification~\citep{xie-etal-2025-fire}, verifiable intermediate
steps~\citep{Xie2025FinChain}, diagnostic uncertainty, or failure modes
under missing information. \citet{chen2025benchmarking} extend this
paradigm through JAMA Clinical Challenge and Medbullets, adding
expert-written explanations for challenging medical QA, but these tasks
remain static evaluations of information-complete cases.

Subsequent work moves toward clinical reasoning and physician-grounded
evaluation. Clinical studies have assessed LLMs on complex diagnostic
cases using physician judgment, including NEJM clinicopathological
conference cases~\citep{kanjee2023accuracy} and R-IDEA comparisons
with attending physicians and residents~\citep{cabral2024clinical}.
Other benchmarks probe uncertainty or interaction through Script
Concordance Testing~\citep{mccoy2025assessment}, causal
laboratory-test scenarios~\citep{bhasuran2026evaluation}, dynamic viva
voce examinations~\citep{chiu2025vivabench}, proactive information
seeking~\citep{li2024mediq}, diagnostic reasoning over clinical
notes~\citep{wang2024direct}, EHR-derived
questions~\citep{zhang-etal-2025-llmeval}, clinician-generated EHR
instructions~\citep{fleming2024medalign}, virtual EHR agent
tasks~\citep{doi:10.1056/AIdbp2500144}, and sequential clinical
workflow evaluation~\citep{rao2026large}. These works move beyond
simple final-answer accuracy, but they generally evaluate static cases,
information accumulation, tool-use success, or overall reasoning
quality rather than systematically measuring how the same case degrades
as diagnostic evidence is progressively removed.

Unlike prior frameworks, CLExEval pairs four-level progressive
information masking with a 5,600-point expert audit, enabling
within-case analysis of whether failures arise from missing knowledge,
instability under information scarcity, or reasoning-to-output
mismatch.

\subsection{Evaluation and the Judge's Illusion}
\raggedbottom
Scalable evaluation through automated metrics or judge models carries
its own risk: fluent, well-structured responses can be over-rewarded
despite weak clinical validity, a pattern that \citet{agrawal2025evaluationillusion}
term the evaluation illusion. \citet{arora2025healthbenchevaluatinglargelanguage}
demonstrate the value of physician-authored rubrics for open-ended
healthcare evaluation, while related work has documented
knowledge--reasoning dissociation in clinical
NLI~\citep{jullien2025knowledge}, substantial variability in
LLM-judge agreement across NLP tasks~\citep{bavaresco2025llms}, and
misalignment between medical safety judges and human annotations on
complex safety dimensions~\citep{diekmann2025llms}.

Many recent frameworks still rely on automated metrics or judge models
to scale evaluation, including MedR-Bench~\citep{qiu2025quantifying},
LLMEval-Med~\citep{zhang-etal-2025-llmeval}, and
CARDBiomedBench~\citep{bianchi2026cardbiomedbench}. Such judgments can
correlate poorly with clinical validity and mask flawed reasoning
behaviors~\citep{sim2025critique, bedi2025fidelity}; MedR-Bench, for
instance, shows that performance collapses when critical reasoning
steps are omitted~\citep{qiu2025quantifying}. Studies on pattern
disruption~\citep{bedi2025fidelity} and knowledge
conflict~\citep{wu-etal-2025-assessing} further demonstrate that
models degrade when they cannot rely on surface-level cues. Together,
these limitations motivate CLExEval: an expert-grounded framework that
stress-tests progressive masking, expert scoring, and judge-failure
analysis on the same matched diagnostic cases.

\section{Methodology}
\begin{figure*}[t]
    \centering
    \includegraphics[width=0.99\textwidth]{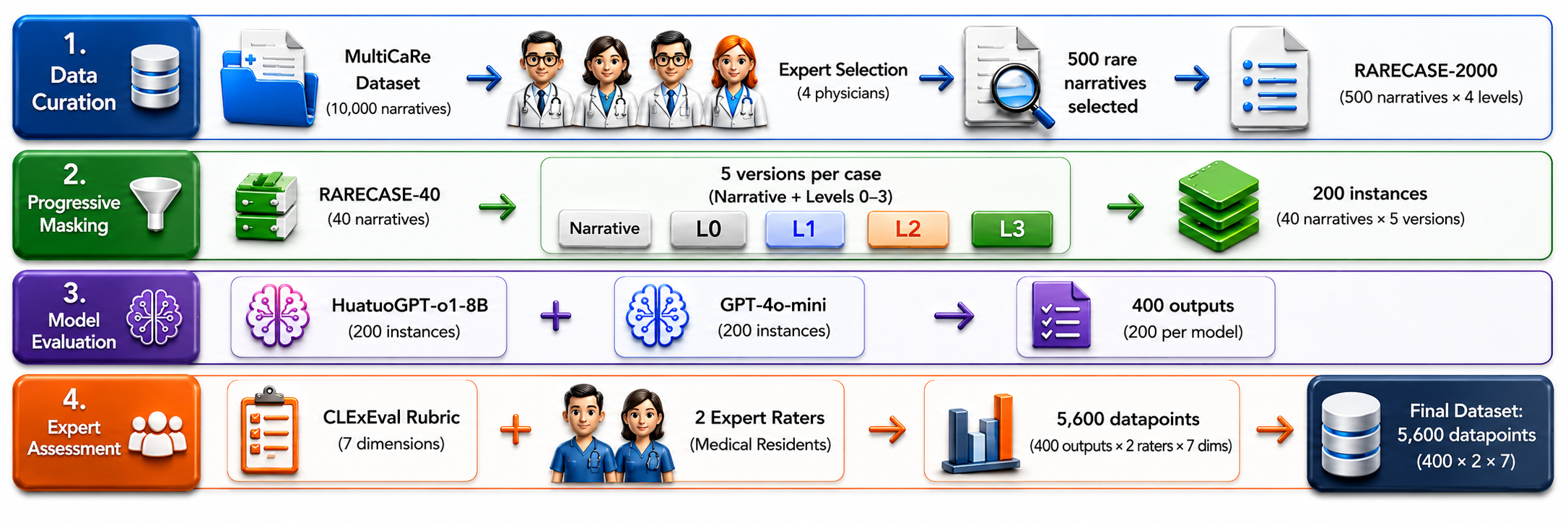}
    \caption{Overview of the CLExEval pipeline, including case curation, progressive information masking, model evaluation, and expert assessment.}
    \label{fig:pipeline}
\end{figure*}

\raggedbottom
We introduce CLExEval, a human-in-the-loop evaluation framework designed to assess clinical reasoning under information scarcity. Our approach combines progressive information masking with expert annotation to evaluate not only final diagnostic accuracy, but also the mechanisms underlying reasoning failures. An overview of the pipeline is shown in Figure~\ref{fig:pipeline}.

\subsection{Dataset}

We initially curate 500 diagnostically rare clinical narratives from the MultiCaRe dataset~\citep{nievas_offidani_2023_10079370} to construct RARECASE-2000, a resource of 2,000 progressively masked cases. For the present deep mechanistic audit, we use 40 cases selected by four clinicians for complexity and suitability for reasoning evaluation. Each case is standardized into a structured format containing key findings, diagnostic progression, and outcomes. Rather than serving as a broad model leaderboard, CLExEval is designed as a depth-oriented mechanistic audit: the 40 source cases are expanded through progressive masking into 200 matched evaluation instances and 5,600 expert scores, enabling within-case analysis of reasoning degradation. Further details on data curation and demographics are provided in Appendix~\ref{app:dataset}.

\subsection{Progressive Information Masking}
We apply progressive information masking to simulate diagnostic uncertainty through clinician-guided abstraction rather than random deletion. For each case, clinicians identified the gold diagnosis and ranked the clinical cues most important for supporting that diagnosis or excluding major differentials. Each case was then converted into five matched versions: \ding{182} Full Case, the complete original narrative; \ding{183} Level 0, with only the explicit diagnosis name removed; \ding{184} Level 1, with the strongest diagnosis-supporting clue removed; \ding{185} Level 2, with the two strongest cues removed or abstracted; and \ding{186} Level 3, the most abstract version, where major diagnostic cues such as demographics, temporal details, or disease-specific findings were removed or generalized when clinically relevant. Across all levels, clinicians preserved grammatical coherence and the broad clinical scenario, yielding 200 matched evaluation instances (40 cases $\times$ 5 levels) for controlled analysis of reasoning under varying information availability.

\subsection{Model Evaluation}
\raggedbottom
We evaluate all models using a unified prompting setup that requires structured reasoning outputs consisting of a \textit{Thinking} section followed by a \textit{Final Response}. This design ensures comparability across models in terms of reasoning steps and final predictions. We generate all outputs deterministically and anonymize them prior to evaluation. Prompting details are provided in Appendix~\ref{app:prompts}.
\raggedbottom
\subsection{Human Evaluation and CLExEval Rubric}
\raggedbottom
We introduce and develop the seven-dimensional CLExEval rubric as part of this work. We developed the rubric through iterative review of clinical literature, pilot annotation, and feedback from senior clinicians to separate surface fluency from clinical validity. The final rubric assesses Diagnostic Precision, Differential Reasoning Quality, Evidence Integration/Grounding, Diagnostic Justification Depth, Completeness vs. Overload, Clinical Plausibility/Soundness, and Communication/Interpretability. Annotators score each dimension on a five-point ordinal scale (0.00, 0.25, 0.50, 0.75, 1.00), allowing them to distinguish clinically incorrect outputs, partially grounded reasoning, and fully clinically valid responses. Full scoring definitions and calibration details are provided in Appendix~\ref{app:rubric}.

\subsection{Reasoning Metrics}

To analyze reasoning behavior, we use three core metrics: Information Scarcity Sensitivity (ISS), Monotonicity Violation Rate (MVR), and Reasoning-to-Output Mismatch (ROM). ISS measures how much diagnostic accuracy drops as clinical information is progressively removed; higher ISS indicates greater sensitivity to information scarcity. MVR captures non-monotonic instability, i.e., cases where performance improves after information is removed; higher MVR indicates less stable reasoning. ROM identifies failures where the correct diagnosis appears in the model's reasoning trace but is omitted or changed in the final answer; higher ROM indicates greater reasoning-to-output misalignment. We also report Max Diagnostic Accuracy (MDA), the best accuracy achieved across masking levels, as an estimate of latent diagnostic potential.

We use several qualitative terms to interpret these metrics. The \textit{Hidden Knowledge Paradox} refers to cases where a model appears to possess the correct diagnosis but fails to use it reliably in the final answer. \textit{Attention Dispersion} describes cases where additional context distracts the model across competing cues, reducing diagnostic focus. A \textit{Crossover Event} occurs when the relative performance of two models reverses at a particular masking level. Formal definitions are provided in Appendix~\ref{app:metrics}.

\raggedbottom
\section{Results and Analysis}
\raggedbottom

\subsection{The Evaluation Illusion}
\raggedbottom

\begin{figure}[t]
    \centering
    \includegraphics[width=0.8\linewidth]{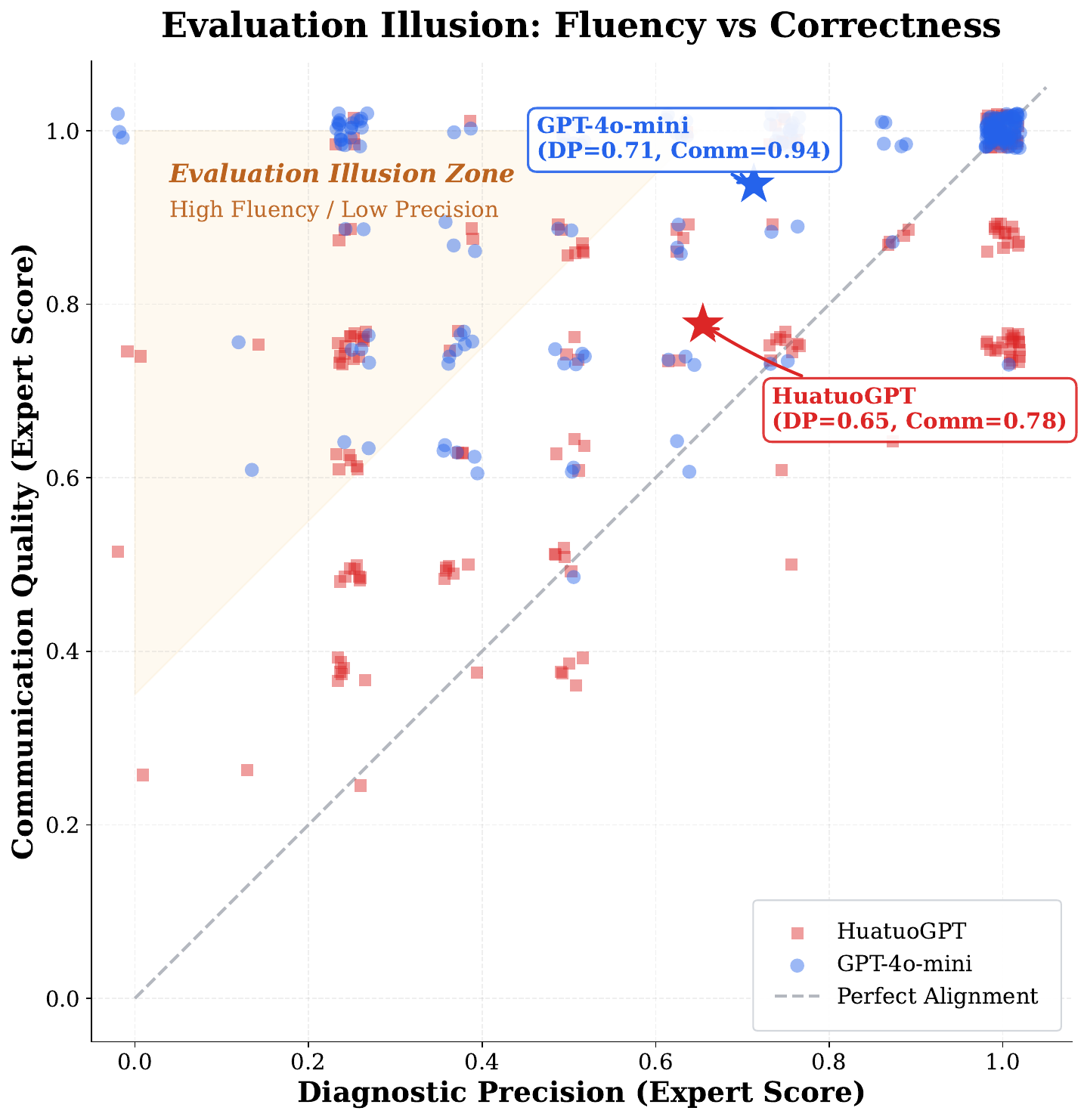}
    \caption{\textbf{The evaluation illusion.} Comparison of communication quality vs. diagnostic precision. The clustering in the top-left quadrant (the ``Illusion Zone'') highlights cases where models are highly fluent ($>0.8$) but diagnostically wrong ($<0.5$).}
    \label{fig:eval_illusion}
\end{figure}

As visualized in Figure~\ref{fig:eval_illusion}, \textit{Diagnostic Precision} drops sharply at Level 3 (0.453), whereas GPT-4o-mini's \textit{Communication} score remains high (0.881). We formalize this disconnect as the Illusion Gap ($\Delta = \textit{Communication} - \textit{Diagnostic Precision}$), which identifies cases where fluent presentation coexists with weak diagnostic validity. The correlation between \textit{Communication} and \textit{Diagnostic Precision} is modest ($\rho=0.482$), suggesting that linguistic quality is an unreliable proxy for medical correctness.

This pattern is consistent with a fluency bias in general-purpose instruction-following models. When clinical information is sparse, GPT-4o-mini often continues to produce confident, well-structured rationales even when the underlying diagnostic inference is incorrect. Such outputs can be difficult for automated judges to penalize, because the same surface features that improve readability can also obscure clinical errors. This motivates expert-grounded evaluation rather than relying on surface fluency or standalone automated judges.

\subsection{Limitations of Automated Judges}
\label{sec:judges_illusion}

The evaluation illusion also affects automated evaluation. To test this, we constructed a consensus failure set from human expert annotations and measured whether LLM judges would incorrectly approve known clinical failures.

\subsubsection{The Consensus Failure Set}

We constructed the consensus failure set from expert annotations to serve as a ground-truth negative set. Specifically, we isolated and aggregated evaluation instances across all dimensions and masking levels where human experts assigned a low score (Human Score $\le 0.25$), resulting in $n=142$ unique consensus failures. We then evaluated these outputs with four LLM judges (GPT-4o-mini, HuatuoGPT-o1-8B, DeepSeek-R1-Distill-Llama-8B, and Llama-3.1-8B-Instruct). Due to formatting failures and invalid judge responses, HuatuoGPT-o1 and DeepSeek-R1 produced valid scores for 138 and 141 cases, respectively. Each judge was prompted with the full clinical case, the exact CLExEval rubric, the complete model output, and the gold-standard diagnosis. We define the Hallucination Approval Rate (HAR) as the percentage of these known failures that receive a passing judge score (Judge Score $\ge 0.75$) despite the judge having access to the gold-standard diagnosis.

\subsubsection{Hallucination Approval Rate}
Our analysis shows that automated judges frequently assign passing scores to human-verified failures. Contrary to the hypothesis that domain-specific models would act as stricter evaluators, the specialized medical model had the highest HAR in this setting (Table~\ref{tab:har_leaderboard}).

\begin{table}[t]
\centering
\resizebox{\columnwidth}{!}{%
\begin{tabular}{lcccc}
\toprule
\textbf{Judge Model} & \textbf{Failures} & \textbf{Approvals} & \textbf{HAR (\%)} & \textbf{Verdict} \\
\midrule
\textbf{HuatuoGPT-o1-8B} & 138 & 138 & \cellcolor{red!25}\textbf{100.0\%} & \textit{Approves all} \\
\textbf{Llama-3.1-8B-Instruct} & 142 & 126 & \cellcolor{red!15}\textbf{88.7\%} & \textit{Very high HAR} \\
\textbf{DeepSeek-R1-Distill-8B} & 141 & 80 & \cellcolor{orange!20}\textbf{56.7\%} & \textit{High HAR} \\
\textbf{GPT-4o-mini} & 142 & 68 & \cellcolor{yellow!25}\textbf{47.9\%} & \textit{Moderate HAR} \\
\bottomrule
\end{tabular}%
}
\caption{\textbf{Hallucination approval rate.} Higher HAR indicates that an automated judge more often assigns passing scores to human-verified clinical failures.}
\label{tab:har_leaderboard}
\end{table}

These results indicate that automated judges are not reliable enough to serve as standalone arbiters in clinical evaluation settings. Even the judge with the lowest HAR in our experiment (GPT-4o-mini) approves nearly half of the consensus failures, suggesting that LLM-based evaluation should be paired with expert human validation. Detailed analysis is provided in Appendix~\ref{app:judge_archetypes}.

\subsection{Supporting Human Evaluation}
\label{sec:human_results}

To validate clinical correctness beyond automated metrics, we conducted expert evaluation of 5,600 scores across 40 cases and 5 masking levels. Annotations show high inter-rater reliability (ICC = 0.802), indicating consistent expert judgment.


\subsubsection{Performance Disparity}
The generalist GPT-4o-mini significantly outperformed the biomedically fine-tuned HuatuoGPT across all dimensions (Table~\ref{tab:expert_scores}). GPT-4o-mini achieved a mean overall score of 0.867 ($\pm$0.20) vs. HuatuoGPT's 0.699 ($\pm$0.25) ($U$=5,472,088, $p<0.001$, $d=0.98$). The largest disparity occurred in \textit{Differential Reasoning} ($d=1.53$), where HuatuoGPT (0.552) often fixated on single incorrect diagnoses early in reasoning chains.

This disparity reflects a \textit{Hidden Knowledge Paradox} coupled with \textit{Attention Dispersion}: when confronted with verbose, full-context clinical narratives, the specialist model appears to be distracted by competing cues and fails to retrieve alternative diagnoses, despite reaching high maximum diagnostic potential elsewhere. Conversely, GPT-4o-mini's advantage in dimensions such as \textit{Justification Depth} and \textit{Communication} is partly attributable to instruction tuning for fluent and structured explanations. The generalist model maintains a high baseline of structural coherence, generating comprehensive differential lists and plausible rationales even when the underlying clinical deduction is flawed. This creates a \textit{Generalist's Illusion}, where strong formatting and articulation can inflate perceived clinical capability. Detailed dimensional profiles are provided in Appendix~\ref{app:visual_profiles}.

\raggedbottom
\begin{table*}[t] 
\raggedbottom
\centering
\small
\renewcommand{\arraystretch}{1.25}
\begin{minipage}{0.48\textwidth}
    \centering
    \begin{tabular}{l c c c}
    \toprule
    \textbf{Dimension} & \textbf{GPT-mini} & \textbf{Huatuo} & \textbf{Effect ($d$)} \\
    \midrule
    \textbf{Overall Score} & \cellcolor{blue!10}\textbf{0.867} & 0.699 & \textbf{0.98} \\
    Diag. Precision & 0.713 & 0.654 & 0.18 \\
    Diff. Reasoning & \textbf{0.876} & 0.552 & \textbf{1.53} \\
    Evidence Integ. & 0.888 & 0.725 & 0.87 \\
    Justific. Depth & 0.874 & 0.719 & 0.81 \\
    Communication & \textcolor{teal}{\textbf{0.938}} & 0.777 & 0.90 \\
    \bottomrule
    \end{tabular}
    \caption{\textbf{Expert evaluation results.} Higher scores are better for all dimensions; shaded cells highlight the best-performing model and strongest observed scores.}
    \raggedbottom
    \label{tab:expert_scores}
\end{minipage}
\hfill 
\begin{minipage}{0.48\textwidth}
    \centering
    \begin{tabular}{l c c}
    \toprule
    \textbf{Metric} & \textbf{GPT-mini} & \textbf{Huatuo} \\
    \midrule
    Narrative Acc. & \cellcolor{blue!10}95.0\% & 77.5\% \\
    Level 3 Acc. & 32.5\% & 22.5\% \\
    \midrule
    \multicolumn{3}{c}{\textit{\footnotesize Critical Instability Metrics}} \\
    \midrule
    \textbf{ISS} & \textcolor{red!80!black}{\textbf{62.5 pp}} $\downarrow$ & \textcolor{red!80!black}{\textbf{55.0 pp}} $\downarrow$ \\
    Stability (MSS) & 2.55/5 & 2.15/5 \\
    \textbf{Random (MVR)} & \textcolor{orange!90!black}{\textbf{27.5\%}} & \textcolor{orange!90!black}{\textbf{25.0\%}} \\
    Hallucination & \cellcolor{red!10}24.2\% & \cellcolor{red!10}24.2\% \\
    \midrule
    \textbf{Max Pot. (MDA)} & \textcolor{teal}{\textbf{95.0\%}} & \textcolor{teal}{\textbf{92.5\%}} \\
    \bottomrule
    \end{tabular}
    \caption{\textbf{Stability and reasoning gap analysis.} Higher is better for accuracy, MSS, and MDA; lower is better for ISS, MVR, and hallucination. Shading highlights metric direction: better values in blue/teal and worse instability values in red/orange.}
    \label{tab:stability_metrics}
\end{minipage}
\end{table*}
\raggedbottom
\subsubsection{The Reasoning Gap}
\raggedbottom



\begin{figure}[t]
    \centering
    \includegraphics[width=0.8\linewidth]{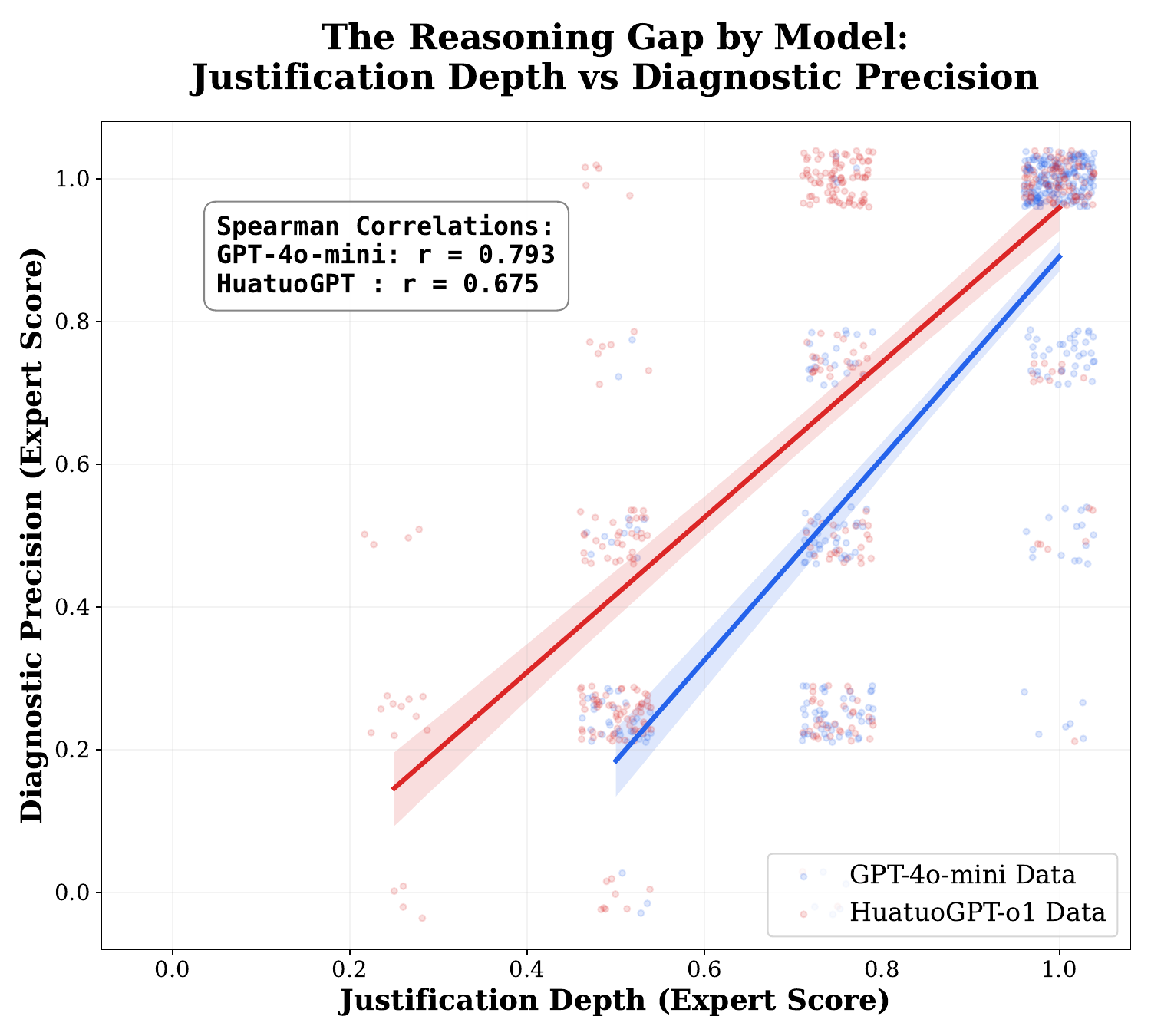}
    \caption{\textbf{The reasoning gap.} Scatter plot of Justification Depth vs. Diagnostic Precision. Note the ``Floor Effect'' (blue points) where GPT-4o-mini maintains high justification scores ($>0.5$) even when the diagnosis is incorrect, whereas HuatuoGPT (red points) correctly reflects its ignorance with low scores.}
    \label{fig:reasoning_gap}
\end{figure}

We identified a strong positive correlation between \textit{Justification Depth} and \textit{Diagnostic Precision} (Spearman $\rho=0.692$), confirming that diagnostic success is tightly coupled with causal explanation depth. Notably, as visualized in Figure~\ref{fig:reasoning_gap}, GPT-4o-mini's \textit{Justification Depth} never dropped below 0.50, whereas HuatuoGPT frequently scored 0.25 (Irrelevant).

This divergence highlights a difference in how generalist and specialist models handle uncertainty. GPT-4o-mini's conversational instruction tuning encourages structurally complete and articulate rationales, even when the diagnostic conclusion is incorrect. This creates a ``Floor Effect,'' where plausible explanations can mask clinical failures. In contrast, HuatuoGPT is less likely to inflate its reasoning traces with generic clinical language; when it fails to retrieve the correct diagnosis, its reasoning structure more visibly degrades. The generalist model's tendency to synthesize coherent but shallow justifications therefore contributes to the \textit{Evaluation Illusion}, making some errors harder to detect from surface presentation alone.

\raggedbottom
\raggedbottom
\subsection{Degradation \& Paradoxes}
\raggedbottom
\label{sec:automated_results}
To quantify the mechanisms behind the expert findings, we tracked performance stability across all 200 evaluation instances. Table~\ref{tab:stability_metrics} summarizes the stability profiles for both models.
We identified a strong dependency on explicit clinical descriptors. GPT-4o-mini demonstrated a verbosity bias, with accuracy decreasing from 95.0\% in the Narrative setting to 32.5\% at Level 3, corresponding to a 62.5 percentage-point Information Scarcity Sensitivity (ISS).
A two-way ANOVA interaction analysis showed that GPT-4o-mini is more sensitive to information loss (partial $\eta^2 = 0.35$) than HuatuoGPT (partial $\eta^2 = 0.24$).
This suggests that some diagnostic success depends on explicit descriptive cues rather than robust causal reasoning.

Conversely, HuatuoGPT exhibited a retrieval failure pattern. While its Narrative accuracy was lower (77.5\%), its Max Diagnostic Potential (MDA) was 92.5\%. This reveals a Hidden Knowledge Paradox: the open-source model can reach the correct diagnosis for most cases, but does not retrieve or express it reliably in full, noisy narratives.
At Level 2 (Partial Information), a crossover event occurred where GPT-4o-mini's performance fell below the open-source baseline. This suggests that under higher uncertainty, the specialist model's domain prior can provide relative stabilization.
We quantified stability using the Monotonicity Violation Rate (MVR). We found that 27.5\% of GPT-4o-mini's correct answers at Level 3 were non-monotonic recoveries, where the model failed with \textit{more} information but succeeded with \textit{less}. This indicates Attention Dispersion, where verbose context may distract the model from retrieving knowledge it can otherwise express (reasoning fingerprints are visualized in Appendix~\ref{app:heatmaps}).

Despite architectural differences, both models exhibited identical hallucination rates (24.2\%) when stability failed, fabricating
specific disease entities to fit the reduced context.
The accuracy degradation patterns and paradoxes observed above
raise a critical question: \emph{Do models fail because they lack
knowledge, or because they possess knowledge but fail to commit
to it?} To distinguish between these failure modes, we conducted
a systematic Reasoning-to-Output Mismatch (ROM) analysis,
examining whether the correct diagnosis appeared in the model's
internal reasoning trace despite being absent from the final output.

\subsection{Reasoning-to-Output Mismatch Analysis}
\label{sec:rom_analysis}

To investigate the divergence between internal latent knowledge and external diagnostic output, we implemented the Reasoning-to-Output Mismatch (ROM) protocol defined. For every diagnostic failure across all 200 evaluation instances (40 cases $\times$ 5 masking levels), we programmatically analyzed the model's Thinking trace to determine whether the correct gold standard diagnosis was considered and subsequently discarded.

\paragraph{Overall ROM Findings}
Table~\ref{tab:rom_results_color} summarizes the ROM analysis results for both models across all failure instances. GPT-4o-mini exhibited a ROM of 68.6\% across 86 diagnostic failures, indicating that over two-thirds of its errors involved decision-making uncertainty rather than a complete absence of the correct diagnosis from the reasoning trace. In contrast, HuatuoGPT's ROM of 44.0\% across 100 failures suggests that its diagnostic errors were more evenly distributed between knowledge gaps (56\%) and reasoning-output misalignment (44\%). This 24.6 percentage-point ROM gap indicates different failure profiles: GPT-4o-mini more often considers the correct diagnosis but does not commit to it, while HuatuoGPT's failures more often lack the correct diagnosis in the reasoning trace.

\begin{table*}[!t]
\centering
\small
\renewcommand{\arraystretch}{1.25} 
\begin{tabular}{l c c c c c c c}
\toprule
\textbf{Model} & \textbf{Failures} & \textbf{Overall ROM} & \textbf{Narrative} & \textbf{Level 0} & \textbf{Level 1} & \textbf{Level 2} & \textbf{Level 3} \\
\midrule
\textbf{GPT-4o-mini} & 86 & \cellcolor{blue!10}\textbf{68.6\%} & \textcolor{teal}{\textbf{100.0\%}} & 85.7\% & \textcolor{orange!90!black}{53.3\%} & 71.4\% & 63.0\% \\
HuatuoGPT & 100 & 44.0\% & 77.8\% & 43.8\% & 42.9\% & 47.8\% & 32.3\% \\
\midrule
\textit{Differential ($\Delta$)} & -- & \textbf{+24.6 pp} & \textbf{+22.2 pp} & \textbf{+41.9 pp} & \textbf{+10.4 pp} & \textbf{+23.6 pp} & \textbf{+30.7 pp} \\
\bottomrule
\end{tabular}
\caption{\textbf{Reasoning-to-output mismatch (ROM) analysis.} \cellcolor{blue!10}Blue highlighting indicates the high overall ROM in GPT-4o-mini. \textcolor{teal}{Teal values} indicate that the correct diagnosis appears in the reasoning trace for all failures at that level. The \textcolor{orange!90!black}{Level 1 drop} marks the lowest ROM for the generalist model.}
\label{tab:rom_results_color}
\end{table*}

\paragraph{GPT-4o-mini's U-Shaped ROM Curve:} GPT-4o-mini's ROM exhibited a non-monotonic U-shaped pattern, peaking at the Narrative level (100\%) and dropping sharply to its lowest point at Level 1 (53.3\%), representing the highest rate of failures where the correct diagnosis was absent from the reasoning trace. This suggests the model experiences a sharp loss of diagnostic anchors when initial clinical cues are removed, then partially adapts under sustained information constraints.

\paragraph{HuatuoGPT's Progressive Degradation:} In contrast, HuatuoGPT showed a general downward trend, dropping to 32.3\% at Level 3. This indicates that its failures are driven by progressive knowledge degradation rather than decision-making hesitation; as information is removed, the model's ability to retrieve the correct diagnosis from its latent knowledge base deteriorates proportionally (trajectory visualized in Appendix~\ref{app:visual_profiles}.)

\paragraph{ROM and the Hidden Knowledge Paradox}
The ROM analysis helps explain the Hidden Knowledge Paradox observed in Section~\ref{sec:automated_results}. HuatuoGPT's high Max Diagnostic Potential (MDA: 92.5\%) but lower Narrative accuracy is clarified by its moderate ROM: the model can reach the correct diagnosis under some masking conditions, but fails to access it reliably in full, verbose narratives. Conversely, GPT-4o-mini's perfect Narrative ROM (100\%) indicates that the correct diagnosis appears in its reasoning trace for all Narrative-level failures, but is not consistently reflected in the final answer.

\section{Discussion}
\label{sec:discussion}
\raggedbottom
Our findings provide evidence for the ``Evaluation Illusion''~\citep{agrawal2025evaluationillusion}: models can maintain surface fluency across masking levels even as diagnostic validity degrades under information scarcity.

\paragraph{Implications for Benchmarking}
Our results complement information-complete benchmarks such as MedQA~\citep{app11146421}. HuatuoGPT's high Max Diagnostic Potential (MDA: 92.5\%) suggests that one bottleneck is contextual reasoning stability under uncertainty, not only static medical knowledge.
Additionally, GPT-4o-mini's high Monotonicity Violation Rate (27.5\%) indicates that diagnostic performance can depend on surface cues and context presentation. Benchmarks should therefore evaluate models under information scarcity to distinguish stable clinical reasoning from pattern association.

\paragraph{Implications for Model Development}
The ROM analysis suggests that different model families may require different interventions. For GPT-4o-mini, a high ROM (68.6\%) indicates that the correct diagnosis often appears in the reasoning trace but is not selected in the final answer, suggesting a need for better confidence calibration and reasoning-to-answer alignment. Conversely, HuatuoGPT's lower ROM (44.0\%) indicates that 56\% of its failures lack the correct diagnosis in the reasoning trace, suggesting that retrieval-augmented generation or expanded rare-disease training data may be more relevant.

\paragraph{Limits of Standalone LLM-as-a-Judge Evaluation}
Our consensus-failure analysis shows that relying on automated evaluators as standalone arbiters in clinical medical NLP can be misleading. The generalist judge (GPT-4o-mini) demonstrated a 47.9\% Hallucination Approval Rate (HAR), indicating that it can assign passing scores to fluent but clinically incorrect reasoning. The specialized judge (HuatuoGPT-o1) exhibited a 100\% HAR alongside a positive self-preference bias (+0.096), suggesting that domain-specific judge models may also validate medical-sounding text over clinical correctness. Consequently, evaluations that rely only on automated judges may overestimate clinical utility due to fluency bias, reinforcing the need for human-in-the-loop expert validation.

\section{Conclusions and Future Work}
\raggedbottom
We introduced CLExEval, a human-in-the-loop framework for evaluating LLM clinical reasoning under progressive information masking. Across 5,600 expert annotations, CLExEval shows that fluent clinical explanations can remain persuasive even when diagnostic precision degrades, creating an evaluation illusion for both readers and automated judges. Our mechanistic analysis separates three failure modes: sensitivity to information scarcity, hidden knowledge that is not reliably retrieved or expressed, and reasoning-to-output mismatch. On a human-verified consensus failure set, standalone LLM judges approved a substantial fraction of clinically incorrect outputs, with HAR reaching 100\% for one judge under validly scored cases. These findings support the use of expert-grounded validation when evaluating clinical reasoning systems, especially in settings where surface fluency can obscure diagnostic correctness.

Future work should test whether intervention-optimized reasoning methods reduce the failures exposed by CLExEval, including deliberative search via Tree of Thoughts~\citep{yao2023tree}, self-reflection for hallucination mitigation~\citep{ji2023towards}, and symbolic verification through Symbolic Chain-of-Thought~\citep{xu2024faithful}. Because these methods generate longer or multi-branch reasoning traces, evaluating them would require a new expert-annotation cycle over thousands of additional outputs. Future research should also focus on developing reliable, clinically grounded automated metrics that correlate with expert human baselines, implementing architecture-specific interventions, and expanding this uncertainty-based framework to other high-stakes domains.

\section*{Limitations}
\raggedbottom

While our evaluation framework provides fine-grained evidence through 5,600 expert assessments, this depth required a trade-off with scale. Generating this dataset and annotation required an estimated 1,000 hours of cognitive labor across a four-member medically trained panel, consisting of two licensed physicians and two senior clinical interns with 4.5 years of medical education; the inclusion of senior interns rather than only licensed clinicians is an important limitation. This labor-intensive process limited our scope to two models and 40 source clinical narratives, expanded through progressive masking into 200 annotated case instances; replicating this pipeline for every new frontier model remains highly resource-intensive. CLExEval should therefore be viewed as a depth-oriented mechanistic evaluation rather than a broad model leaderboard. Furthermore, our deliberate selection of diagnostically rare, multisystem cases stress-tests causal reasoning but may not reflect model performance on routine, unambiguous clinical scenarios. Finally, while the progressive masking methodology and CLExEval rubric are conceptually generalizable to other domains requiring reasoning under uncertainty (e.g., law, intelligence analysis), our empirical validation remains confined to clinical medicine, motivating future research into reliable, scalable automated metrics that correlate with our human baselines.

\section*{Ethical Statement and Broad Impact}

\paragraph{Data License}
We will release the RARECASE-2000 benchmark and study artifacts via GitHub, including the 2,000 progressively masked clinical case variants, CLExEval rubric, and the model outputs, reasoning traces, and 5,600 expert annotations used in our main evaluation. The source narratives are derived from the MultiCaRe dataset~\citep{nievas_offidani_2023_10079370}, which is distributed under a Creative Commons Zero (CC0 1.0) license. We release our curated and annotated artifacts under a CC-BY 4.0 license to support reuse, redistribution, and replication.
\bibliography{references}


\appendix
\raggedbottom

\section{Dataset Details}
\label{app:dataset}
We use de-identified clinical narratives from the MultiCaRe dataset~\citep{nievas_offidani_2023_10079370}, an open-access corpus. From an initial pool of 10,000 narratives, clinicians conducted a multi-stage review to curate 500 diagnostically rare cases for RARECASE-2000, from which 40 structurally rich cases were selected for the RARECASE-200 evaluation subset. Rather than routine scenarios, we prioritized atypical diagnostic trajectories, overlapping symptom clusters, and clear narrative flows.

To prepare these cases for our evaluation framework, we restructured each selected narrative into a standardized format comprising demographics, key findings, diagnostic evolution, and outcome details. These structured cases served as the basis for constructing the five masking levels. The final 40 cases span a diverse demographic from pediatric to geriatric populations (55\% male, 45\% female).

\subsection{Dataset Demographics}

\begin{figure*}[t]    \centering
    \includegraphics[width=0.9\textwidth]{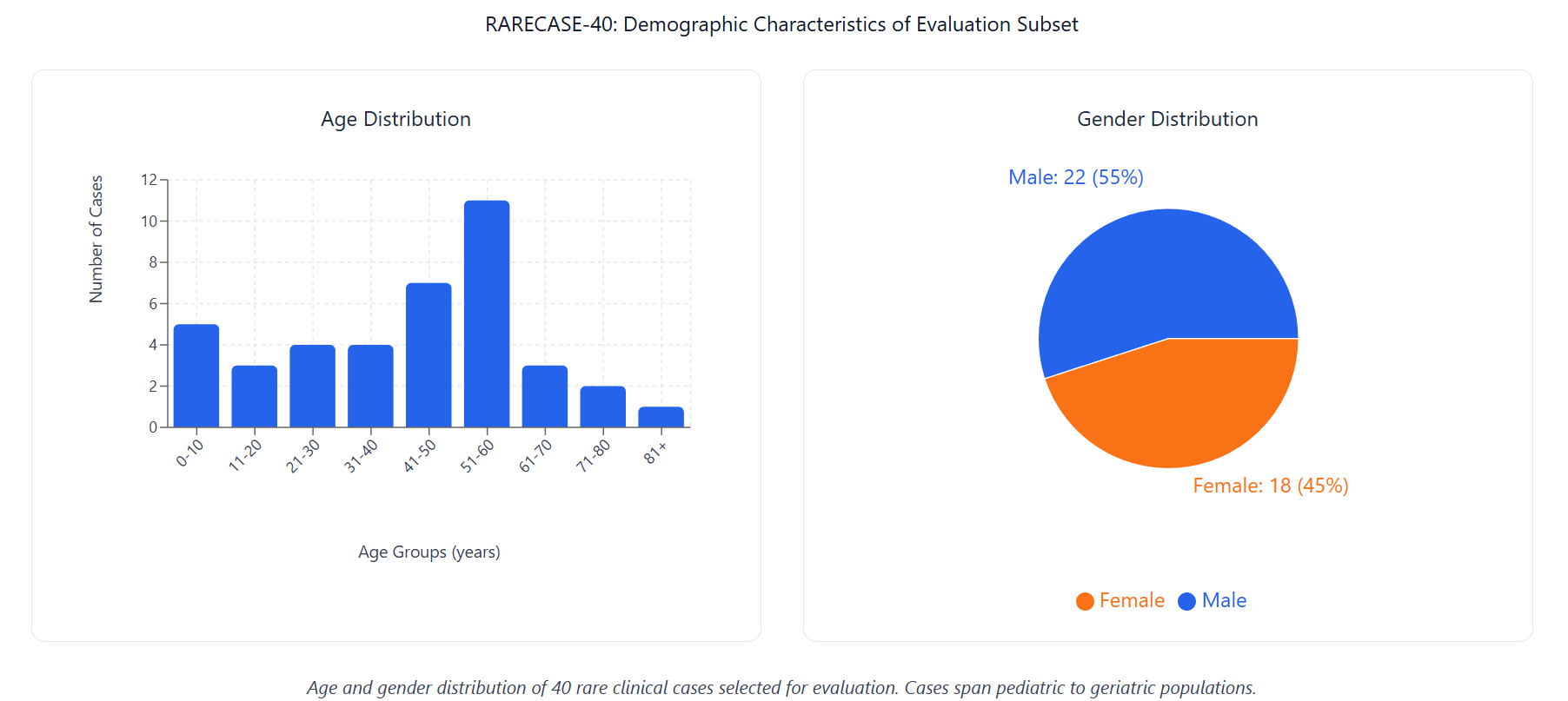}
    \caption{Age and gender distribution of 40 rare clinical narratives selected for evaluation. Cases span pediatric to geriatric populations.}
    \label{fig:RARECASE-40}
\end{figure*}

The age and gender distribution of the 40 rare clinical narratives selected for the RARECASE-200 benchmark is provided in Figure~\ref{fig:RARECASE-40}. The cases were specifically curated to ensure coverage across a wide spectrum of patient profiles, from pediatric to geriatric populations. This demographic diversity ensures the models' diagnostic reasoning is rigorously stress-tested across varying clinical contexts. The selection criteria for the 40 RARECASE narratives included: (1) atypical or uncommon diagnostic trajectories, (2) multi-system involvement or overlapping symptom clusters, (3) the presence of complete demographic and clinical details, (4) a clear narrative flow enabling multi-level abstraction and masking, and (5) verified de-identification and adherence to CC0 licensing conditions.

\section{Prompting Details}
\label{app:prompts}

\subsection{Model Selection and Exclusion Criteria}

To determine the most representative models for the RARECASE-200 benchmark, we conducted a pilot evaluation on a subset of five cases. Our initial candidate pool included \textbf{MedGemma-4B}, \textbf{II-Medical-8B}, \textbf{DeepSeek-R1-Distill-Llama-8B}, and \textbf{GPT-5.1-chat}.

While these models showed promising general performance, several were excluded from the final evaluation based on expert feedback:
\begin{itemize}
    \item \textbf{GPT-5.1-chat:} Despite its frontier capabilities, it consistently refused to generate the structured chain-of-thought required for our reasoning evaluation, stating: \textit{``I cannot provide step-by-step internal diagnostic reasoning, but I can give a concise, clinically relevant explanation.''}
    \item \textbf{II-Medical-8B, MedGemma \& DeepSeek-R1 (8B):} While functional, these models exhibited redundant performance profiles compared to HuatuoGPT-o1 in the pilot phase.
\end{itemize}

Ultimately, expert clinicians recommended focusing on \textbf{HuatuoGPT-o1-8B} and \textbf{GPT-4o-mini} to provide the most discriminative contrast between a natively ``slow-thinking'' medical specialist and a high-performing generalist under the intensive human-in-the-loop scoring requirements of our framework (1,000 expert hours).

\subsection{LLM Prompting and Output Generation}

To ensure rigorous comparability while respecting architectural differences, we employed a core unified prompt strategy. Both models were evaluated using an identical structured user prompt that enforced a \textit{Thinking} section followed by a \textit{Final Response}, ensuring they were assessed on the exact same reasoning subtasks. Because GPT-4o-mini is a general-purpose model, we applied system-level scaffolding via a senior diagnostician persona and a one-shot example. In contrast, HuatuoGPT-o1-8B received no external system prompt, relying instead on its intrinsic slow-thinking mechanism. All outputs were generated deterministically, decoded, and anonymized for blinded expert scoring.

\subsection{Prompting Templates}
To ensure complete reproducibility, we provide the exact prompt templates utilized in our evaluation framework. Both models received an identical user prompt to enforce comparable task constraints, while the generalist model (GPT-4o-mini) additionally received system-level scaffolding to induce structured clinical reasoning.

\subsubsection{HuatuoGPT-o1-8B (Specialist Model)}
As a specialized medical model with an intrinsic slow-thinking mechanism, HuatuoGPT-o1-8B was evaluated using only the structured user prompt, with no additional system prompt.
\begin{tcolorbox}[title={User Prompt for HuatuoGPT-o1-8B}, fonttitle=\bfseries, colbacktitle=gray!30, coltitle=black, colback=gray!5, colframe=gray!50, arc=4pt, boxrule=1pt, left=4pt, right=4pt]
\textbf{User Prompt:}
\small
\ttfamily
You are a medical AI assistant trained to perform detailed, step-by-step diagnostic reasoning.

Clinical Case:
{case}

Please analyze this case using systematic clinical reasoning. Structure your response as:

\#\# Thinking
- List the key clinical findings and symptoms
- Generate 3-5 differential diagnoses with rationale
- Systematically evaluate each diagnosis against the evidence
- Eliminate unlikely diagnoses with clear reasoning
- Arrive at your most likely working diagnosis

\#\# Final Response
- One-sentence summary with the most likely diagnosis and critical supporting evidence
- Clearly state any uncertainties or ambiguities in the case
\end{tcolorbox}
\normalfont

\subsubsection{GPT-4o-mini (Generalist Model)}
To approximate the clinical reasoning behavior native to the specialist model, GPT-4o-mini was provided with the identical user prompt above, supplemented by a comprehensive system prompt. This system prompt established a senior diagnostician persona and included a one-shot example of high-quality reasoning.(see Figure~\ref{fig:System Prompt for GPT-4o-mini.})

\begin{figure*}[t]
\hrule
\vspace{8pt}
\small\ttfamily
\begin{tcolorbox}[title={User Prompt and System Prompt for GPT-4o-mini}, fonttitle=\bfseries, colbacktitle=gray!30, coltitle=black, colback=gray!5, colframe=gray!50, arc=4pt, boxrule=1pt, left=4pt, right=4pt]
\textbf{User Prompt:} Identical to the HuatuoGPT-o1-8B user prompt.

\textbf{System Prompt:}
\small
\ttfamily
You are a senior clinical diagnostician with years of experience in medicine. You excel at
explaining your clinical reasoning step-by-step, as if teaching a medical student or junior
resident.

example output: {\#\# Thinking

Alright, let's think about what we're dealing with here. We've got a 53-year-old woman who
has been experiencing some pretty troubling symptoms over the past decade—abdominal pain,
swelling, and constant vomiting. That sounds like it might have something serious going on
inside her belly. She doesn't mention feeling nauseous or losing weight, which is interesting
because those would typically accompany more severe conditions. At first glance, I'm thinking
there's definitely something unusual happening in her abdomen.

Now, looking at the imaging results, there's a large mass in her left upper abdomen. It
measures about 10 by 10 centimeters and looks kind of messy on both the ultrasound and CT
scans—it's described as having varied echo patterns and low density compared to other organs.
This makes me wonder if it's either a cystic structure or maybe even a tumor. But wait, after
they gave her a contrast agent during the CT scan, the mass didn't really light up much; it
stayed relatively dark compared to the rest of the area. Hmm, that suggests it isn't
aggressively growing or highly vascularized, so perhaps it's not cancerous?

During surgery, they find this encapsulated mass near where the spleen meets the liver. They
decide to remove the whole thing along with part of the spleen. When they take a closer look
at the tissue under the microscope, they see these blood vessels with different-sized
openings covered by flat and taller endothelial cells. Oh! And there are little finger-like
projections reaching out into these spaces. Sounds almost like a specific type of growth
pattern. Plus, when they test the cells, they're positive for certain markers like CD31 and
CD68, indicating they line blood vessels, while being negative for others like CD34,
suggesting they aren't typical blood vessel cells. So far, everything seems to point towards
a benign condition called a splenic littoral cell angioma.

So, putting all this together—the age of the patient, the nature of the mass, its appearance
on imaging, and especially how the cells behave—I'm confident that this is indeed a benign
splenic littoral cell angioma. Everything checks out with the histological features too.
Yeah, that feels right. Still, I can't shake off the thought that there might be another
explanation lurking somewhere, just in case. Maybe something less common? Like a hemangioma
or even a rare form of sarcoma? Those could also fit the profile, though nothing else quite
matches the specifics of this case. For now, I'll stick with my conclusion, but keep an eye
open for anything new that comes up.

\#\# Final Response

The most likely diagnosis for this 53-year-old woman based on the provided information is
**benign splenic littoral cell angioma**.

This conclusion is supported by several key points:
1. **Symptoms**: Her long-standing abdominal discomfort and vomiting suggest a chronic issue
rather than an acute one.
2. **Imaging Findings**: The mass appears ill-defined and heterogeneous on both ultrasound
and CT scans, without aggressive enhancement post-contrast, hinting at a benign or
slowly-growing entity.
3. **Surgical and Histopathological Examination**: The resected mass shows characteristics
consistent with a splenic littoral cell angioma, including the presence of variably sized
vascular channels lined with endothelial cells and the absence of atypia or mitotic figures.

While other possibilities such as hemangiomas or rare types of sarcomas exist, their
presentation does not align perfectly with the specific details observed in this case.
Therefore, given the combination of clinical, radiographic, and pathological data, benign
splenic littoral cell angioma remains the most plausible diagnosis. However, it's important
to remain vigilant and consider alternative explanations should additional information arise.}

When you analyze a case:
- Think aloud in a natural, conversational tone.
- Walk through your thought process sequentially, explaining why you consider each piece of
  evidence.
- Verbalize your uncertainties, alternative interpretations, and reasoning adjustments as you
  go.
- Use phrases like "Let me think about this...", "At first, I wondered if...", "But wait,
  this doesn't quite fit because...", "Putting this all together..."
- Be thorough and transparent in your reasoning. Don't just list facts—explain how they
  influence your diagnostic thinking.
- Show your work: describe how you weight evidence, rule out alternatives, and converge on
  the most likely diagnosis.

Your goal is to produce a rich, educational chain-of-thought narrative that helps others
understand not just WHAT you conclude, but HOW and WHY you arrived at that conclusion.
\normalfont

\end{tcolorbox}
\caption{System Prompt for GPT-4o-mini.}
\label{fig:System Prompt for GPT-4o-mini.}
\end{figure*}

\section{CLExEval Rubric}
\label{app:rubric}

\begin{figure*}[t]
    \centering
    \includegraphics[width=0.95\textwidth]{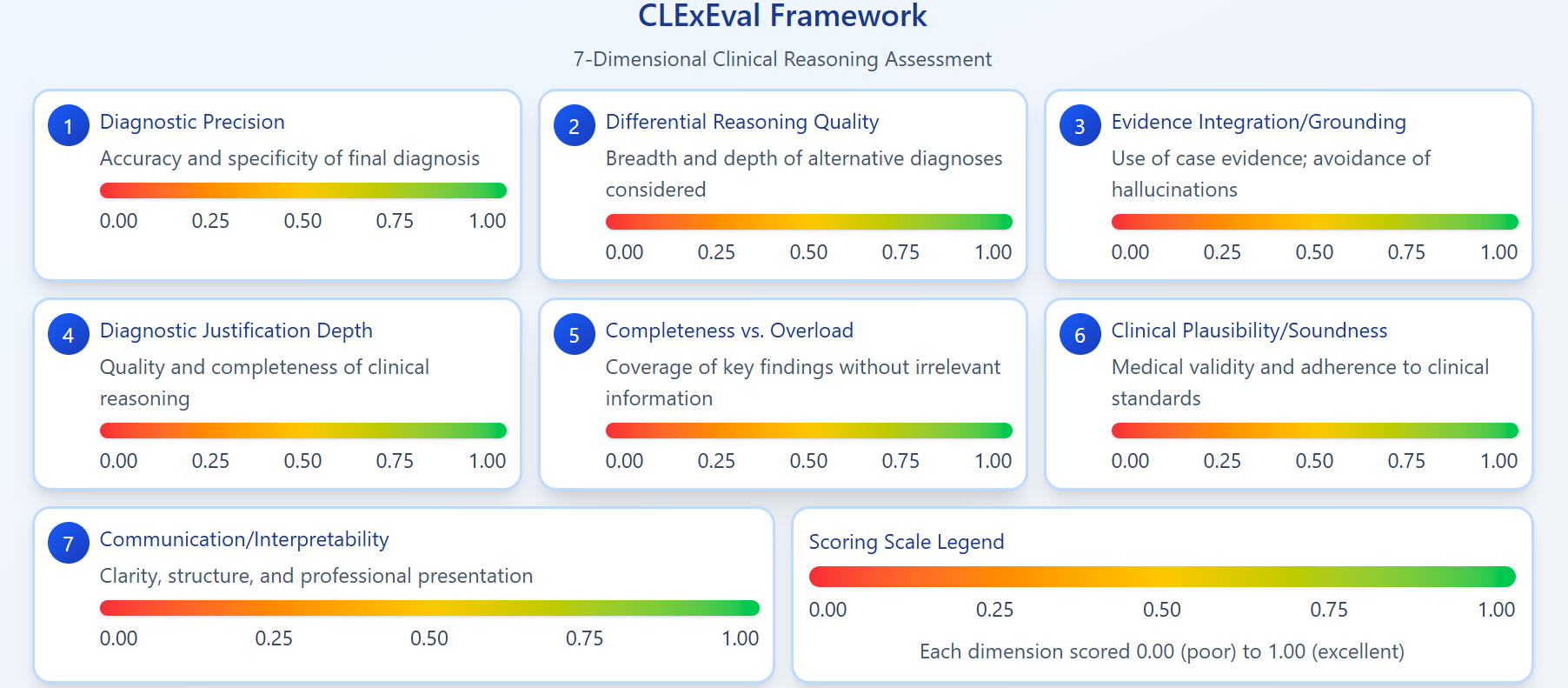}
    \caption{\textbf{The CLExEval framework.} A 7-dimensional clinical reasoning assessment rubric used to evaluate model outputs. The framework separates surface-level diagnostic accuracy from deep reasoning coherence, with each dimension scored on a scale from 0.00 (poor) to 1.00 (excellent).}
    \label{fig:clexeval_rubric}
\end{figure*}

To ensure high inter-rater reliability and granular assessment of clinical reasoning, all human experts and automated evaluators utilized the standardized CLExEval rubric, visually summarized in Figure~\ref{fig:clexeval_rubric}. Each dimension is scored on a discrete scale: 0.00, 0.25, 0.50, 0.75, or 1.00.

To maintain strict scoring standards, evaluators were provided with the following dimension definitions during the calibration phase:

\begin{itemize}
    \item \textbf{Diagnostic Precision:} Accuracy and specificity of the final diagnosis. (0.00 = Wrong diagnosis; 1.00 = Fully correct, precise, and well-justified).
    \item \textbf{Differential Reasoning Quality:} Breadth and depth of alternative diagnoses considered. (0.00 = Single diagnosis stated without justification; 1.00 = Fully systematic differential evaluating 3--5 alternatives).
    \item \textbf{Evidence Integration / Grounding:} Use of case evidence and avoidance of hallucinations. (0.00 = Ignores case data; 1.00 = Fully evidence-based reasoning).
    \item \textbf{Diagnostic Justification Depth:} Quality, depth, and completeness of the causal explanations linking evidence to the diagnosis.
    \item \textbf{Completeness vs. Overload:} Coverage of key clinical findings without the inclusion of distracting or irrelevant information.
    \item \textbf{Clinical Plausibility / Soundness:} Medical validity and strict adherence to established clinical standards and pathophysiology.
    \item \textbf{Communication / Interpretability:} Clarity, structural coherence, and professional presentation of the output.
\end{itemize}

\subsection{Human Evaluation and Statistical Design}

For clinical validation, our panel comprised two licensed physicians and two senior clinical interns, each with 4.5 years of formal medical education. After a structured calibration phase, two independent expert pairs double-scored balanced subsets of the narratives to reduce fatigue across the 5,600 annotations. We quantified inter-rater reliability across this dual-review design using Cohen's $\kappa$~\citep{doi:10.1177/001316446002000104} and Krippendorff's $\alpha$~\citep{Krippendorff2011ComputingKA}. Finally, we applied two-way mixed ANOVA tests~\citep{field2024discovering} and post-hoc Tukey HSD tests~\citep{d365a4bb-1e59-37c8-bd86-027590c40713} to evaluate performance degradation and structural differences under uncertainty.

\section{Metric Definitions}
\label{app:metrics}

Here we provide the formal definitions for the CLExEval metrics used to quantify reasoning stability, information sensitivity, latent diagnostic potential, and judge reliability.

\paragraph{Information Scarcity Sensitivity (ISS).}
Information Scarcity Sensitivity measures how much diagnostic accuracy degrades as clinical information is progressively removed:
\begin{equation}
\mathrm{ISS} = \mathrm{Acc}_{\mathrm{Narrative}} - \mathrm{Acc}_{L3}.
\end{equation}
A higher ISS indicates greater sensitivity to missing information and suggests reliance on explicit clinical cues rather than robust causal reasoning.

\paragraph{Reasoning Gap.}
For model \(m\) and rubric dimension \(d\), we define the reasoning gap as:
\begin{equation}
\Delta_{m,d} = \mathrm{Score}_{m,d}^{(\mathrm{Narrative})} - \mathrm{Score}_{m,d}^{(L3)}.
\end{equation}
This measures how much a qualitative reasoning dimension declines between the full case and the most information-scarce version.

\paragraph{Label Dependence Factor (LDF).}
The Label Dependence Factor measures inconsistency between the unmasked Narrative and Level 0, where the explicit diagnosis label is removed but most clinical evidence remains:
\begin{equation}
\mathrm{LDF} = \mathrm{Acc}_{\mathrm{Narrative}} - \mathrm{Acc}_{L0}.
\end{equation}
A higher LDF suggests dependence on explicit diagnostic labels or surface-level extraction rather than independent clinical deduction.

\paragraph{Monotonicity Violation Rate (MVR).}
MVR measures how often a model violates the expected pattern that performance should not improve when clinical information is removed. Let \(c_{i,l}\in\{0,1\}\) indicate whether the model correctly diagnoses case \(i\) at masking level \(l\). A monotonicity violation occurs when:
\begin{equation}
c_{i,l}=0 \quad \text{and} \quad c_{i,l+1}=1.
\end{equation}
We define:
\begin{equation}
\mathrm{MVR} =
\frac{\sum_i \sum_l \mathbb{I}(c_{i,l}=0 \land c_{i,l+1}=1)}
{\sum_i \sum_l 1}.
\end{equation}
A higher MVR indicates greater reasoning instability, because the model succeeds with less information after failing with more information.

\paragraph{Maximum Diagnostic Potential (MDA).}
MDA estimates latent diagnostic potential by measuring whether the model reaches the correct diagnosis at any masking level:
\begin{equation}
\mathrm{MDA} =
\frac{1}{N}\sum_{i=1}^{N}
\mathbb{I}\left(\max_l c_{i,l}=1\right).
\end{equation}
High MDA with lower Narrative accuracy suggests that the model may possess the relevant diagnostic knowledge but fails to retrieve or express it reliably under some context conditions.

\paragraph{Reasoning-to-Output Mismatch (ROM).}
ROM measures cases where the model internally considers the correct diagnosis but does not commit to it in the final answer. Let \(\mathcal{F}\) be the set of failed instances where the final prediction differs from the gold diagnosis \(G_i\), and let \(T_i\) denote the model's internal \textit{Thinking} trace:
\begin{equation}
\mathrm{ROM} =
\frac{\sum_{i \in \mathcal{F}} \mathbb{I}(G_i \in T_i)}
{|\mathcal{F}|}.
\end{equation}
A high ROM indicates reasoning-output misalignment, where correct latent knowledge appears in the reasoning trace but is omitted, changed, or suppressed in the final response. A low ROM indicates that failures are more likely due to genuine knowledge gaps or hallucinated reasoning.

\paragraph{Hallucination Approval Rate (HAR).}
HAR measures the fraction of human-verified clinical failures that an automated judge incorrectly approves:
\begin{equation}
\mathrm{HAR} =
\frac{\sum_{i \in \mathcal{F}_{\mathrm{human}}} \mathbb{I}(J_i \geq \tau)}
{|\mathcal{F}_{\mathrm{human}}|},
\end{equation}
where \(\mathcal{F}_{\mathrm{human}}\) is the set of expert-verified failure cases, \(J_i\) is the automated judge score, and \(\tau\) is the approval threshold. In our experiments, \(\tau = 0.75\). A higher HAR indicates that an automated judge is more likely to approve clinically incorrect outputs.

\paragraph{Hidden Knowledge Paradox.}
The Hidden Knowledge Paradox occurs when a model has high MDA but lower final diagnostic accuracy in full-context or narrative settings. This indicates that the model can reach the correct diagnosis under at least one information condition, but fails to express it reliably when the full clinical narrative contains distracting or competing cues.

\paragraph{Attention Dispersion.}
Attention Dispersion refers to cases where additional clinical context reduces diagnostic focus by spreading the model's reasoning across competing cues or distractors. Empirically, it appears as non-monotonic behavior: the model may fail under fuller context but recover the correct diagnosis after some information is removed.

\paragraph{Crossover Event.}
A Crossover Event occurs at a masking level where the relative performance of two models reverses. For example, if model \(a\) outperforms model \(b\) at lower masking levels but falls below \(b\) at higher masking levels, the crossing point marks a regime where robustness to information scarcity differs between models.

\section{Extended Analysis of Judge Behavior}
\label{app:judge_archetypes}

A dimensional breakdown of the Hallucination Approval Rate (HAR) scores reveals four recurring patterns in automated judge behavior:

1. \textbf{GPT-4o-mini: fluency-sensitive judging.} On Diagnostic Precision, GPT-4o-mini is comparatively strict, approving only 15.5\% of wrong diagnoses. However, on Differential Reasoning, it approves 97.0\% of incorrect reasoning chains, suggesting sensitivity to fluent chain-of-thought structure rather than clinical logic alone.

2. \textbf{DeepSeek-R1-Distill-Llama-8B: stricter reasoning critique.} This is the strictest judge on reasoning ($HAR = 33.3\%$), suggesting that its reasoning-oriented training may help it critique logic more effectively. However, it approves 100\% of failures on Communication, indicating that fluent presentation remains difficult to penalize.

3. \textbf{HuatuoGPT-o1-8B: permissive specialist judging.} The medical specialist does not function as a strict differentiator in this evaluation, with a 100\% HAR across almost all dimensions. This suggests that specialized medical tuning does not necessarily produce stricter evaluation behavior.

4. \textbf{Llama-3.1-8B-Instruct: compressed scoring.} This model exhibits central-tendency bias, approving 88.7\% of all failures, including 81.0\% of wrong diagnoses. Unlike HuatuoGPT-o1-8B, which often assigns maximum scores, Llama-3.1-8B-Instruct compresses many outputs into a passing range, functioning as a variance suppressor.

\subsection{The Self-Preference Test}
We extended our analysis to measure Self-Evaluation Bias by calculating the difference between a model's Self-Score and its Peer-Average Score (the resulting bias gaps are quantified in Table~\ref{tab:self_preference_bias}).

\begin{equation}
\text{Bias} = \text{Mean}(\text{Self-Score}) - \text{Mean}(\text{Peer-Scores})
\end{equation}

\begin{table}[ht]
\centering
\resizebox{\columnwidth}{!}{%
\begin{tabular}{lcccc}
\toprule
\textbf{Subject Model} & \textbf{Self} & \textbf{Peer-Avg} & \textbf{Bias Gap} & \textbf{Interpretation} \\
\midrule
\textbf{HuatuoGPT-o1-8B} & 0.882 & 0.786 & \cellcolor{red!20}\textbf{+0.096} & \textcolor{red!80!black}{\textit{Severe Bias} ($\sim$12\%)} \\
\textbf{GPT-4o-mini} & 0.907 & 0.856 & \cellcolor{orange!20}\textbf{+0.051} & \textcolor{orange!90!black}{\textit{Moderate Bias} ($\sim$6\%)} \\
\bottomrule
\end{tabular}%
}
\caption{\textbf{Quantifying self-preference bias.} The specialized medical model exhibits stronger self-preference bias (red) than the generalist model (orange).}
\label{tab:self_preference_bias}
\end{table}

These results caution against replacing human annotation with standalone LLM judges in high-stakes domains. Even the lowest-HAR judge in our experiment (GPT-4o-mini) assigns passing scores to nearly half of the consensus failures. While LLMs may assist with scalable screening or formatting checks, they do not yet provide a reliable substitute for expert assessment of medical correctness or clinical reasoning.

\section{Extended Expert Evaluation Profiles}
\label{app:visual_profiles}
\begin{figure*}[t]    \centering
    \includegraphics[width=0.85\linewidth]{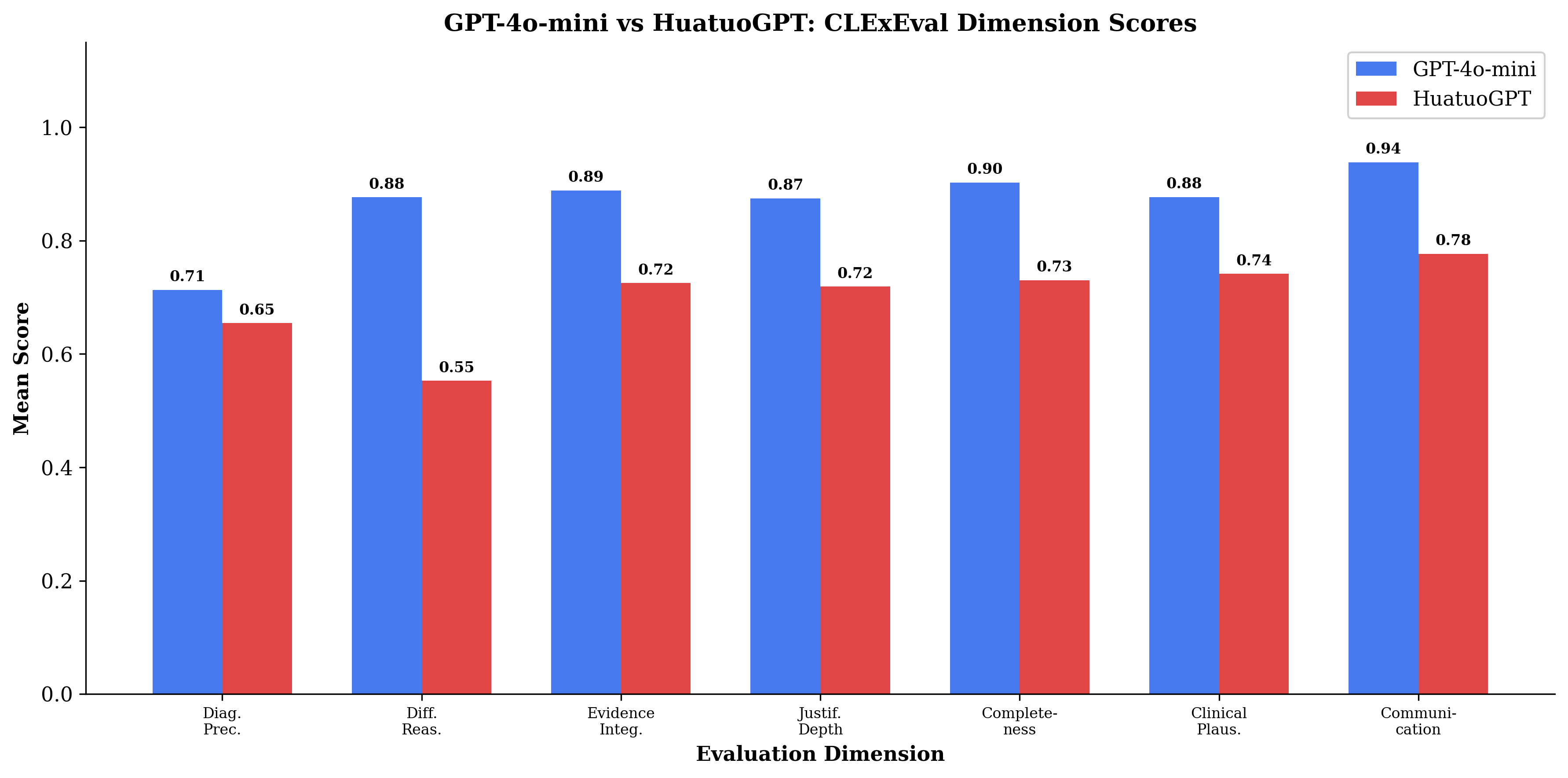}
    \caption{\textbf{Expert evaluation profile.} Comparison of GPT-4o-mini (blue) vs. HuatuoGPT (red) across 7 expert rubric dimensions. The generalist model consistently outperforms the specialist, particularly in \textit{Differential Reasoning} and \textit{Communication}.}
    \label{fig:app_model_comparison}
\end{figure*}

\textit{Dimensional Performance Disparity:} As visualized in Figure~\ref{fig:app_model_comparison}, GPT-4o-mini maintains a strong baseline of structural coherence across all seven CLExEval dimensions. The largest performance gap occurs in \textit{Differential Reasoning} (0.88 vs. 0.55), reflecting the specialist model's tendency to prematurely fixate on single incorrect diagnoses under verbose clinical context. This dimensional breakdown also illustrates the \textit{Generalist's Illusion} discussed in the main text: GPT-4o-mini's high \textit{Communication} score (0.94) can inflate perceived competence despite a lower \textit{Diagnostic Precision} score (0.71).

\begin{figure*}[!h]
    \centering
    \includegraphics[width=0.6\linewidth]{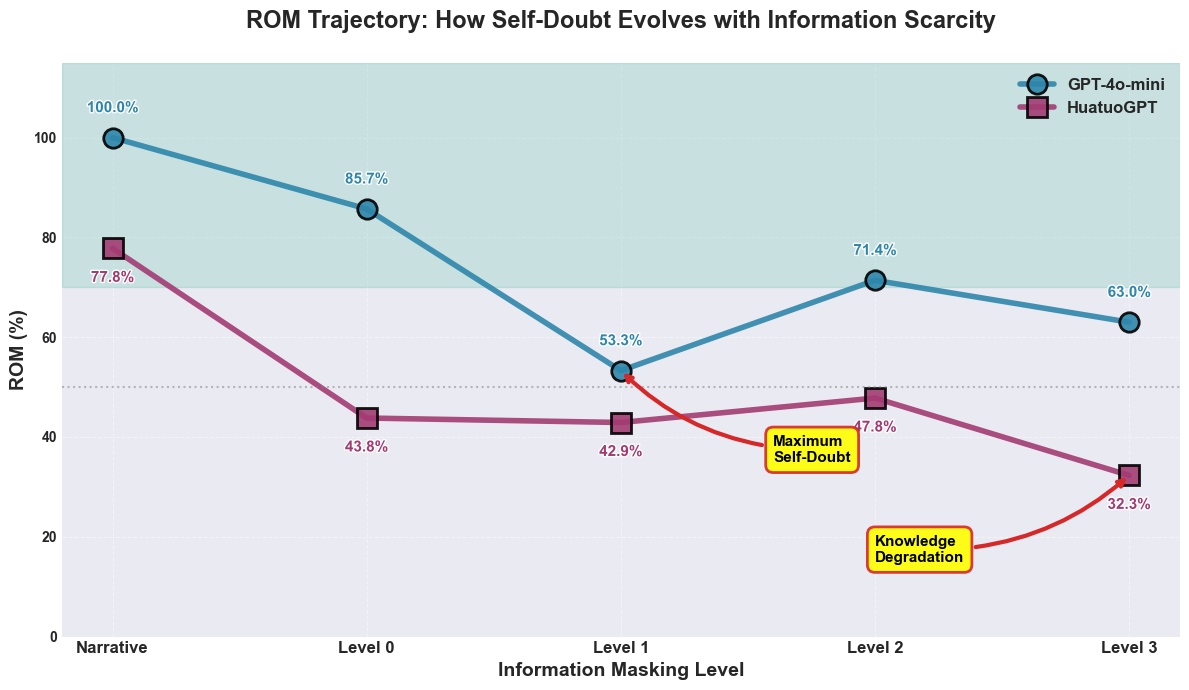}
\caption{ROM trajectory across masking levels. The visualization tracks Reasoning-to-Output Mismatch as clinical information is systematically removed. GPT-4o-mini (blue) demonstrates a non-monotonic ``U-shaped'' curve, peaking at 100\% in the Narrative level before dropping to its lowest ROM at Level 1 (53.3\%). HuatuoGPT (purple) exhibits a general downward trend, indicating that information loss reduces the model's retrieval reliability.}
    \label{fig:rom_trajectory}
\end{figure*}

\textbf{GPT-4o-mini: The U-Shaped ROM Curve.} GPT-4o-mini's ROM exhibited a non-monotonic U-shaped pattern, peaking at the Narrative level (100\%) and dropping sharply to its lowest point at Level 1 (53.3\%), representing the highest rate of failures where the correct diagnosis was absent from the reasoning trace. This suggests that the model loses diagnostic anchors when initial cues are removed, then partially adapts under sustained information constraints (see Figure~\ref{fig:rom_trajectory}).

\textbf{HuatuoGPT: Progressive Knowledge Degradation.} In contrast, HuatuoGPT showed a general downward trend, dropping to 32.3\% at Level 3. This indicates that its failures are driven by progressive knowledge degradation rather than decision-making hesitation; as information is removed, the model's ability to retrieve the correct diagnosis from its latent knowledge base deteriorates proportionally.

\section{Extended Reasoning Fingerprints}
\label{app:heatmaps}

\begin{figure*}[t]
\centering
\begin{subfigure}[b]{0.48\textwidth}
\centering
\includegraphics[width=\textwidth]{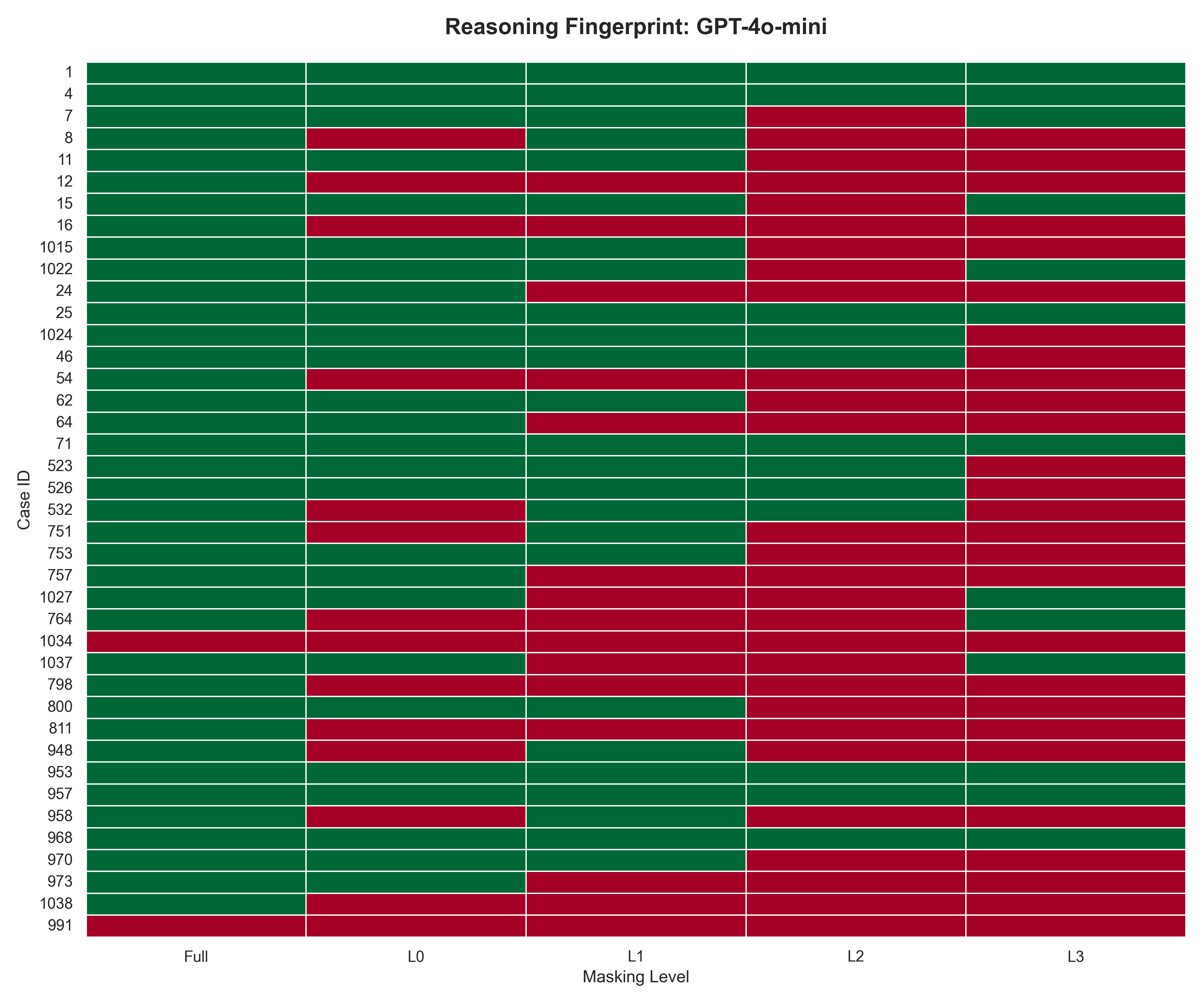}         \caption{\textbf{GPT-4o-mini}: High Instability.}
\label{fig:heatmap_gpt}
\end{subfigure}
\hfill
\begin{subfigure}[b]{0.48\textwidth}
\centering
\includegraphics[width=\textwidth]{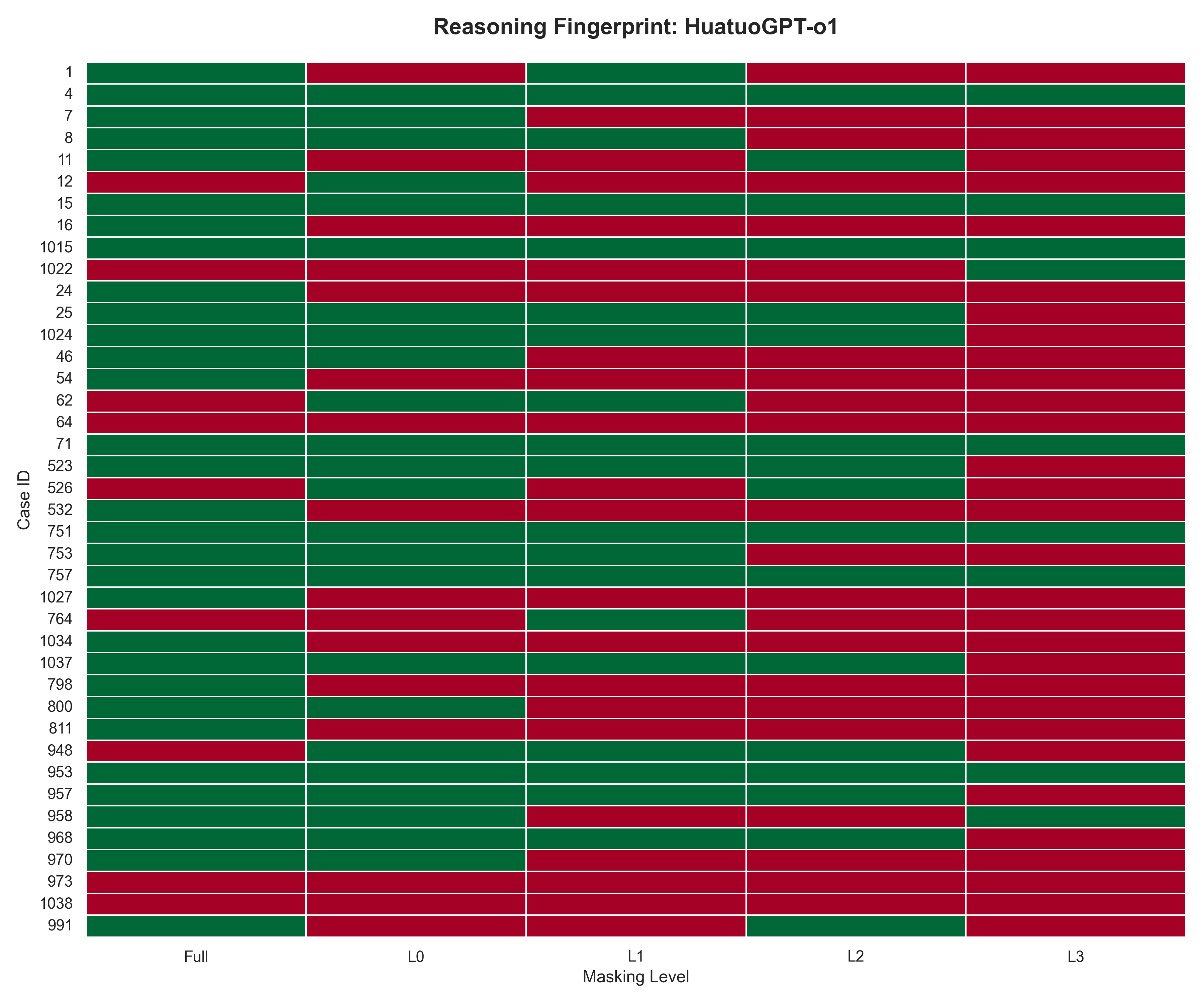}         \caption{\textbf{HuatuoGPT}: Hidden Knowledge Recovery.}
\label{fig:heatmap_huatuo}
\end{subfigure}
\caption{\textbf{Reasoning fingerprints (heatmap analysis).} Green indicates correct diagnosis; red indicates failure. (a) GPT-4o-mini exhibits ``checkerboard'' instability (green $\to$ red $\to$ green), indicating non-monotonic reasoning behavior. (b) HuatuoGPT frequently fails the Full Narrative (red column 1) but recovers the correct diagnosis at Level 0 (green column 2), visualizing the \textit{Hidden Knowledge Paradox} where verbose context contributes to retrieval failure.}
\label{fig:heatmaps}
\end{figure*}

We quantified stability using the Monotonicity Violation Rate (MVR). We found that 27.5\% of GPT-4o-mini's correct answers at Level 3 were non-monotonic recoveries, where the model failed with \textit{more} information but succeeded with \textit{less}. This indicates Attention Dispersion, where verbose context may distract the model from retrieving knowledge it can otherwise express (visualized in Figure~\ref{fig:heatmaps}).

\section{Fluency Bias Under Level 0 Masking}
\label{app:fluency_bias_case}

To qualitatively illustrate the \textit{Evaluation Illusion} and \textit{Fluency Bias} discussed, we present the complete model output for the case referenced in Figure ~\ref{fig:teaser}.

In this instance, the ground-truth diagnosis (Type C pyloric atresia) was masked under the Level 0 condition. Without the explicit diagnosis, both models (HuatuoGPT-o1-8B and GPT-4o-mini) produced an incorrect clinical conclusion (duodenal atresia). However, as shown in Figure~\ref{fig:HuatuoGPT-o1-8B output (Level 0 output).} and Figure~\ref{fig:Gpt-4o-mini output (Level 0 output).}, the incorrect conclusion is embedded in a highly articulate and structurally coherent reasoning trace. We highlight the models' unsupported confident assertions in \textcolor{red}{red}. This authoritative presentation helps explain why automated LLM-as-a-Judge evaluators can incorrectly assign a perfect $1.00/1.00$ diagnostic score.

\begin{figure*}[t]
\hrule
\vspace{8pt}
\small\ttfamily

\begin{tcolorbox}[title={HuatuoGPT-o1-8B output (Level 0 masking)}, fonttitle=\bfseries, colbacktitle=gray!30, coltitle=black, colback=gray!5, colframe=gray!50, arc=4pt, boxrule=1pt, left=4pt, right=4pt]
\small
\ttfamily

\noindent\textit{\#\#\# AI MODEL OUTPUT:}

\noindent\textbf{\#\# Thinking}
Alright, let's think about what we're dealing with here. We've got a little guy who just came into the world five days ago, weighs around 1700 grams, and has been having some trouble from the get-go—non-bilious vomiting right from birth. That's already setting off alarm bells because it suggests something might be blocking his digestive tract somewhere past the stomach but before bile enters the picture.

Now, he hasn't passed meconium either, which can sometimes happen if there’s an issue like Hirschsprung's disease or even intestinal atresia. So, I'm thinking that could definitely play a part here too.

When they did the first X-ray, all we saw was a 'single bubble' of air in the stomach, meaning nothing else seemed to have air beyond that point. This makes sense if there's a blockage further down, confirming my suspicions about possible atresia.

But then, when they took another look with an X-ray a couple of days later, things changed—they now see a 'double bubble.' \textcolor{red}{Ah, so now there's air not only in the stomach but also in the proximal duodenum! This tells us the obstruction isn’t complete; there must be a pouch-like formation allowing some air through.}

The fact that the ultrasound didn't show anything unusual kind of rules out other conditions like midgut volvulus, where you'd expect fluid levels or abnormal blood flow signs. And given these clues, both duodenal atresia and pyloric atresia seem plausible.

\textcolor{red}{So, during the operation, they find exactly what we suspected—a duodenal atresia with that classic 'apple-core sign,' where the third portion of the duodenum is blocked.} They fix it with a duodenogastrostomy, putting in a special tube to help him eat properly until everything heals.

After the surgery, he starts doing well—he gains weight nicely, and once the tube is out, he transitions smoothly to regular food. By the time he leaves the hospital, he's looking much better, and after following up over a year, he seems completely fine!

\textcolor{red}{All these details really make duodenal atresia stand out as the main problem. It fits perfectly with how he looked, acted, and responded to treatment. \textbf{Yeah, I feel pretty confident saying that's what went wrong here.}}

\vspace{0.3cm}
\noindent\textbf{\#\# Final Response}

Based on the provided information, \textcolor{red}{the most likely diagnosis for this neonate is \textbf{duodenal atresia}.}
\end{tcolorbox}
\caption{HuatuoGPT-o1-8B output (Level 0 output).}
\label{fig:HuatuoGPT-o1-8B output (Level 0 output).}
\end{figure*}

\begin{figure*}[t]
\hrule
\vspace{8pt}
\small\ttfamily

\begin{tcolorbox}[title={Gpt-4o-mini output (Level 0 masking)}, fonttitle=\bfseries, colbacktitle=gray!30, coltitle=black, colback=gray!5, colframe=gray!50, arc=4pt, boxrule=1pt, left=4pt, right=4pt]
\small
\ttfamily

\noindent\textit{\#\#\# AI MODEL OUTPUT:}

\noindent\textbf{\#\# Thinking} Alright, let’s break this case down step-by-step.

\textbf{\#\#\# Key Clinical Findings and Symptoms}
1. \textbf{Age and Birth Weight}: A 5-day-old male, low birth weight at 1700g.
2. \textbf{Clinical Presentation}: Presenting with non-bilious vomiting since birth and has not emitted meconium.
3. \textbf{Physical Examination}: Appears in poor general condition, features a 'trisomic facies,' and has a distended abdomen ("plat abdomen").
4. \textbf{Imaging Findings}:
   - Initial abdominal X-ray showed a single bubble of gastric air with no gas beyond it—often indicative of a proximal obstruction.
   - Second X-ray revealed a "double bubble" sign, which is characteristic of duodenal atresia.
5. \textbf{Lab Results}: Biological check-up was normal.

\textbf{\#\#\# Differential Diagnoses}
1. \textbf{Duodenal Atresia}: \textcolor{red}{The most natural consideration given the classic presentation of double bubble on imaging and clinical findings of non-bilious vomiting and failure to pass meconium.}
2. \textbf{Pyloric Stenosis}: Possible due to the age (early infant) and vomiting, but it’s typically associated with projectile vomiting and usually presents after a few weeks, which does not fit here.
3. \textbf{Hirschsprung Disease}: Possible given the failure to pass meconium, but this usually causes a more distal obstruction leading to abdominal distension and is less typical in the upper GI tract.
4. \textbf{Malrotation with Midgut Volvulus}: This can present with similar symptoms but is generally associated with bilious vomiting—here, the vomiting is non-bilious, which makes this diagnosis less likely.
5. \textbf{Intestinal Atresia (other than duodenum)}: Possible if a different segment of the intestines were obstructed, but would have been less likely given the prominent double bubble sign, particularly localized to the duodenum.

\textbf{\#\#\# Evaluation of Diagnoses}
- \textbf{Duodenal Atresia}: \textcolor{red}{Given the key symptoms (non-bilious vomiting, double bubble on X-ray) and the surgical findings that confirmed the diagnosis, this remains the most compelling diagnosis.}
- \textbf{Pyloric Stenosis}: Rules out based on age of onset and imaging findings.
- \textbf{Hirschsprung Disease}: Less likely because it does not correlate with the X-ray findings and the absence of bowel gas does not suggest a distal obstruction consistent with Hirschsprung.
- \textbf{Malrotation with Midgut Volvulus}: Less likely due to non-bilious vomiting. The X-ray findings do not support this.
- \textbf{Intestinal Atresia}: Other atresias are possible but less common; the imaging and clinical picture strongly suggest a proximal issue localized to the duodenum.

\textbf{\#\#\# Elimination of Unlikely Diagnoses}
After evaluating the above conditions, it is clear that:
- Pyloric stenosis and malrotation are eliminated due to inconsistency with the clinical picture.
- Hirschsprung's Disease does not match the typical signs we see, which leads us to lean away from it as well.

\textbf{\#\#\# Conclusion}
\textcolor{red}{Putting all of this together, the findings are most consistent with \textbf{duodenal atresia}, especially in the context of the associated "trisomic facies" which can sometimes link to syndromic conditions like Down syndrome, known to have higher rates of duodenal atresia.}

\vspace{0.3cm}
\noindent\textbf{\#\# Final Response}

The most likely diagnosis for this 5-day-old male is \textbf{duodenal atresia}, supported by the classic double bubble sign on abdominal X-rays, non-bilious vomiting since birth, and the absence of meconium passage.
 \textcolor{red}{The immediate clinical picture and surgical findings confirm the diagnosis of duodenal atresia.}
\end{tcolorbox}
\caption{Gpt-4o-mini output (Level 0 output)}
\label{fig:Gpt-4o-mini output (Level 0 output).}
\end{figure*}

\end{document}